\documentclass[10pt,a4paper,logo]{paper}

\usepackage[all]{hypcap}

\usepackage[authoryear, round]{natbib}
\usepackage{xspace}

\usepackage{nicefrac}       %
\usepackage{multirow}
\usepackage{cleveref}
\usepackage{dsfont}
\usepackage{makecell}
\usepackage{subcaption}
\usepackage{bbm}
\usepackage{wrapfig}
\usepackage{mathtools,xparse}
\usepackage{array}
\usepackage{arydshln}
\usepackage{scalerel}
\usepackage{tablefootnote}
\usepackage{tocloft}
\usepackage{adjustbox}
\usepackage{setspace}
\usepackage[most,skins,theorems]{tcolorbox}

\newboolean{showsection}
\setboolean{showsection}{true}
\makeatletter
\@namedef{ver@everyshi.sty}{}
\makeatother

\definecolor{custom_green}{rgb}{0.0, 0.5, 0.0}
\definecolor{custom_red}{rgb}{1.0, 0.01, 0.24}
\definecolor{custom_blue}{HTML}{C9DAF7}
\definecolor{custom_purple}{HTML}{D9D1E9}
\definecolor{title_blue}{HTML}{204899}
\definecolor{cite_blue}{HTML}{044dc1}
\definecolor{cite_purple}{HTML}{7406a7}

\usepackage{amsmath,amsfonts,bm}

\def\eqref#1{equation~\ref{#1}}

\def\1{\bm{1}}

\def\vzero{{\bm{0}}}

\def\va{{\bm{a}}}

\def\vs{{\bm{s}}}

\def\vu{{\bm{u}}}

\def\mA{{\bm{A}}}

\def\mI{{\bm{I}}}

\def\mV{{\bm{V}}}

\DeclareMathAlphabet{\mathsfit}{\encodingdefault}{\sfdefault}{m}{sl}
\SetMathAlphabet{\mathsfit}{bold}{\encodingdefault}{\sfdefault}{bx}{n}

\def\gL{{\mathcal{L}}}

\def\gN{{\mathcal{N}}}

\newcommand{\E}{\mathbb{E}}

\makeatletter
\def\mathcolor#1#{\@mathcolor{#1}}
\def\@mathcolor#1#2#3{%
  \protect\leavevmode
  \begingroup
    \color#1{#2}#3%
  \endgroup
}
\makeatother

\Crefformat{equation}{#2Eq.\;(#1)#3}
\crefformat{equation}{#2Eq.\;(#1)#3}
\Crefformat{figure}{#2Figure #1#3}
\crefformat{figure}{#2Figure #1#3}
\Crefformat{assumption}{#2Assumption #1#3}
\Crefname{assumption}{Assumption}{Assumptions}

\usepackage{crossreftools}
\makeatletter
\renewcommand\footnoterule{%
  \kern 15\p@
  \hrule \@width 2in \kern 2.6\p@
  \vspace{4pt}
}
\makeatother

\usepackage{url}

\usepackage{xspace}
\usepackage{dsfont}
\usepackage{afterpage}

\usepackage{enumitem}
\usepackage{booktabs}
\usepackage{caption} %
\usepackage{multirow} %
\usepackage{enumitem}
\usepackage{colortbl} %
\usepackage{bbm}
\usepackage{algorithm}
\usepackage{algpseudocode}                  %
\usepackage{setspace}                       %
\usepackage{subcaption} %
\usepackage[most]{tcolorbox}
\usepackage{fvextra}
\usepackage{xcolor}

\usepackage{mathtools} %
\usepackage{bm} %
\usepackage{soul} %
\usepackage{nicefrac}
\usepackage{csquotes}

\usepackage{wrapfig}
\usepackage{cleveref}

\usepackage{makecell}
\usepackage{placeins}
\usepackage{float}

\newcommand{\eg}{\emph{e.g.}} 
\newcommand{\ie}{\emph{i.e.}}

\definecolor{cornellred}{rgb}{0.7, 0.11, 0.11}
\definecolor{cadmiumgreen}{rgb}{0.0, 0.42, 0.24}
\definecolor{aliceblue}{rgb}{0.91, 0.94, 0.97}
\definecolor{darkblue}{rgb}{0.83, 0.89, 0.97}
\definecolor{Red7}{rgb}{0.941, 0.243, 0.243}
\definecolor{Green7}{RGB}{55, 178, 77}
\definecolor{Blue9}{rgb}{0.098,0.3,0.9}

\usepackage{pifont}%
\definecolor{checkblue}{RGB}{45, 85, 160}
\definecolor{crossred}{RGB}{180, 50, 50}

\definecolor{deltadown}{RGB}{190,70,70}
\definecolor{deltaup}{RGB}{40,110,170}

\sethlcolor{aliceblue}

\hypersetup{
  linkcolor = cornellred,
  citecolor  = cadmiumgreen,
  colorlinks = true,
  urlcolor = violet
}

\newcommand{\sname}{\textsc{PaceVLA}\xspace}

\title{PaceVLA: Adaptive Temporal Action Quantization in Vision-Language-Action Models}

\reportnumber{}
\paperurl{https://pacevla.github.io/}

\author[1,\footnotesize{\textbf{*}}]{Hojin~Jeon}
\author[1,\footnotesize{\textbf{*}}]{Jimin~Lee}
\author[3]{Taekwan~Kim}
\author[1]{Nayoung~Oh}
\author[1,2]{Taeyoung~Kim}
\author[1,2]{Myungkyu~Koo}
\author[4]{Jinmo~Koo}
\author[2]{Joonwoo~Ahn}
\author[1,2]{Jinwoo~Shin}

\affil[1]{KAIST}
\affil[2]{RLWRLD}
\affil[3]{Korea University}
\affil[4]{Seoul National University}

\correspondingauthor{}

\begin{abstract}

Fast and reliable manipulation is crucial for practical robot deployment. 
Although Vision-Language-Action (VLA) models have recently shown remarkable progress as generalist robot policies, their execution often remains temporally inefficient. 
A key source of this inefficiency is that policies learn fine-grained temporal patterns from teleoperated demonstrations, requiring many low-level action steps. 
\emph{Temporal action quantization} offers a simple way to accelerate execution by aggregating consecutive low-level actions into temporally coarser actions. However, applying action quantization without considering when fine-grained control is required introduces a speed–accuracy trade-off: accelerated actions may efficiently reduce redundant steps in some phases, while discarding critical motion details in precision-sensitive interactions.
To address this, we propose \sname{}, a unified framework for adaptive temporal execution in VLAs that integrates two coupled components: (1) mode-specific action decoders built on a shared action expert, supporting multiple temporal granularities and prediction horizons, and (2) a quantizability-guided mode selection scheme which decides whether coarse execution is appropriate, and a context-conditioned router that selects an execution mode at each inference step. 
Experiments across three multi-task simulation benchmarks and six real-world tasks show that \sname{} improves execution efficiency while maintaining task success rate across multiple embodiments, including long-horizon and dexterous manipulation. The \href{https://pacevla.github.io/}{project page} is available.
\end{abstract}

\begin{document}

\maketitle
\newlength{\origtextfloatsep}\setlength{\origtextfloatsep}{\textfloatsep}
\newlength{\origfloatsep}\setlength{\origfloatsep}{\floatsep}
\newlength{\origintextsep}\setlength{\origintextsep}{\intextsep}
\setlength{\parskip}{0.51\baselineskip}
\addtolength{\textfloatsep}{1pt}
\addtolength{\floatsep}{1pt}
\addtolength{\intextsep}{1pt}
\titlespacing*{\section}{0pc}{3.1ex plus4pt minus3pt}{5pt}
\titlespacing*{\subsection}{0pc}{2.58ex plus3pt minus2pt}{2pt}

\begin{figure}[ht]
\centering
\begin{subfigure}[b]{0.67\linewidth}
  \includegraphics[width=\linewidth]{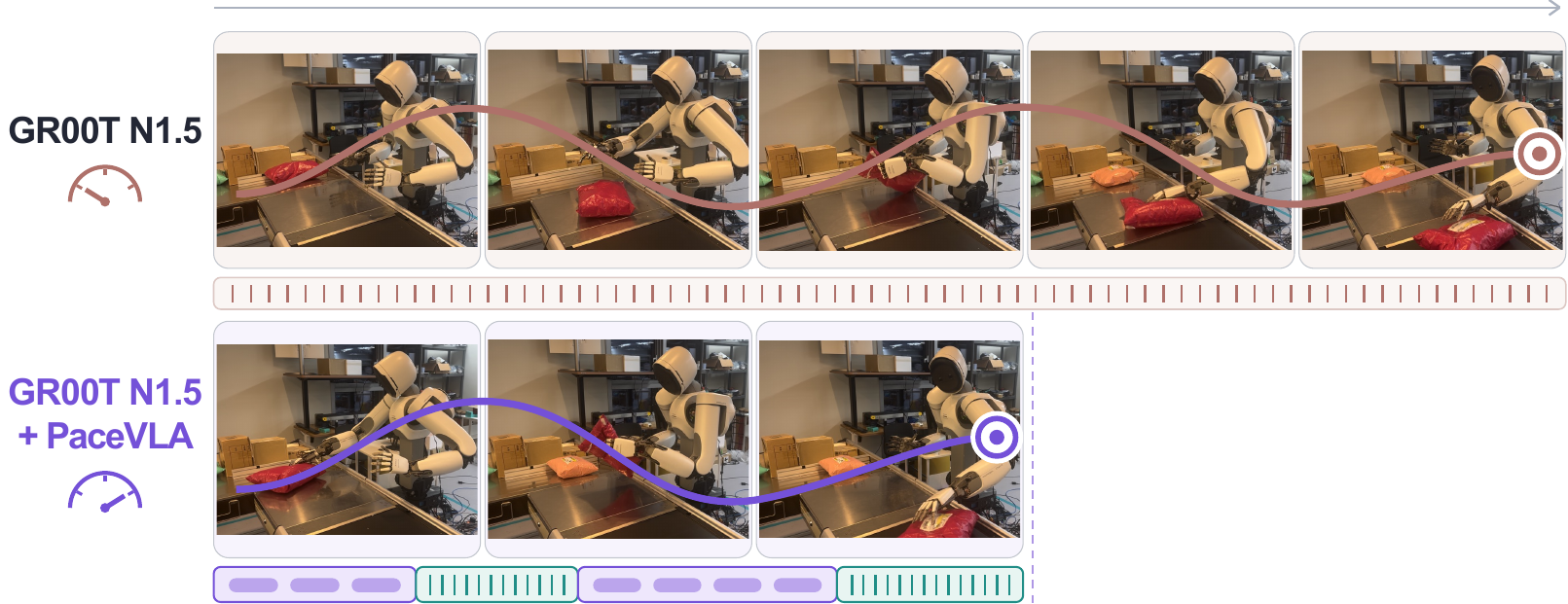}
  \caption{Accelerated execution on the real-robot conveyor task.}
  \label{fig:teaser-demo}
  \vspace{-0.05cm}
\end{subfigure}\hfill
\begin{subfigure}[b]{0.32\linewidth}
  \includegraphics[width=\linewidth]{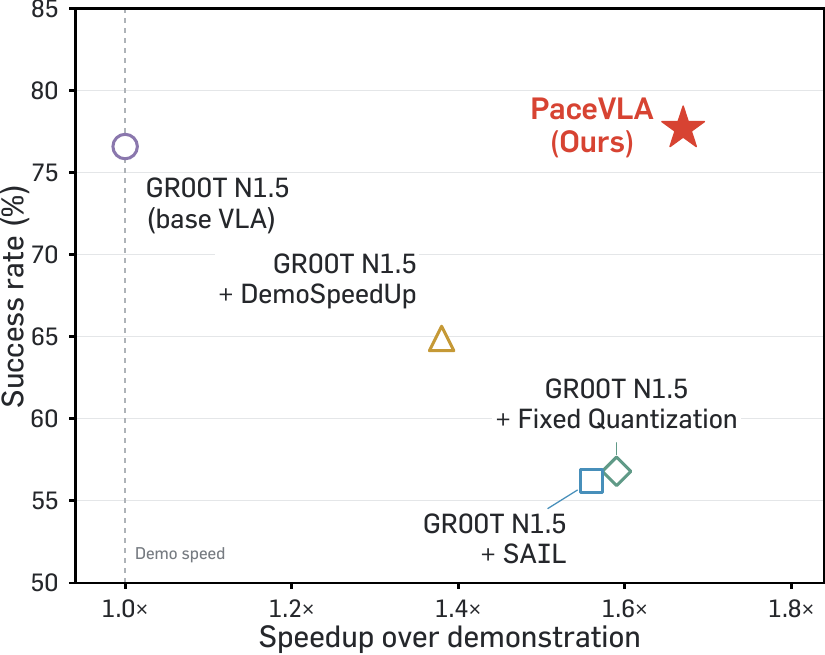}
  \caption{Speed--accuracy trade-off.}
  \label{fig:teaser-pareto}
\end{subfigure}
\caption{\textbf{Adaptive temporal execution with \sname{}.}
\textbf{(a)}~Conveyor package handling task on the humanoid ALLEX: \sname{} selects a temporal granularity per phase.
\textbf{(b)}~Success rate vs.\ speedup over the base VLA, averaged over six real-world tasks.}
\label{fig:teaser}
\end{figure}

\section{Introduction}

Vision-Language-Action models (VLAs)~\citep{zitkovich2023rt, kim2024openvla, intelligence2025pi_, pertsch2025fast, nvidia2025gr00t} have emerged as a promising framework for generalist robot policies, achieving strong performance through imitation learning from diverse robot demonstrations~\citep{walke2023bridgedata, o2024open, khazatsky2024droid}, many of which are collected through human teleoperation~\citep{zhao2023learning,wu2024gello}.
However, teleoperated demonstrations often exhibit conservative behavior: human-in-the-loop control is typically performed through limited-rate interfaces, and human operators adopt cautious motions to avoid failures during data collection~\citep{mandlekar2018roboturk, mandlekar2021matters}.
Since imitation learning trains policies to reproduce the demonstrated trajectories, VLAs inherit these patterns and execute tasks through many slow action steps even in phases where faster motion is physically feasible. 

Recent works have explored accelerating robot policies beyond demonstration speed, addressing two complementary questions: \emph{where} to accelerate and \emph{how coarsely} to execute actions.
DemoSpeedup~\citep{guo2025demospeedup} segments per-frame action entropy with a proxy policy and trains a policy on the resulting accelerated trajectories.
SAIL~\citep{arachchige2025sail} combines high-fidelity tracking with switching between preset slow and fast speeds based on motion complexity and gripper events.
A particularly simple mechanism is \emph{temporal action quantization}~\citep{figure2025helixlogistics, tang2025vlash}, which merges consecutive actions into coarser ones.
It fits naturally with chunk-based VLAs since each predicted action chunk can be executed at a chosen granularity.

Despite their promise, these approaches have two limitations. 
First, the acceleration decision is either fixed into the training data, using the action entropy of a fine-tuned proxy policy as an implicit indicator~\citep{guo2025demospeedup}, 
or applied at execution time to the output of a policy trained at demonstration speed, switching between preset speeds according to kinematic cues~\citep{arachchige2025sail}.
Both signals are statistics of the actions, yet how much precision a phase demands is determined by the robot's interaction with the scene rather than by the motion itself.
Second, existing temporal action quantization applies a fixed granularity across the entire trajectory, inducing a speed--accuracy trade-off: coarse actions suffice during free-space motion but discard trajectory details required in precision-critical phases such as grasping and contact-rich interaction.
Moreover, granularity is closely coupled with the prediction horizon: phases that tolerate coarse execution can commit to longer predictions, whereas precision-critical phases benefit from shorter horizons with more frequent re-planning.
We argue that both the temporal granularity and the prediction horizon of action execution should be treated as a policy decision, adapted to the precision each phase requires.

To this end, we propose \sname{}, a unified framework for adaptive temporal action execution in VLAs.
\sname{} equips a shared action expert with mode-specific action decoders that support different temporal granularities and prediction horizons. At each inference step, a quantization judge and a router jointly select an execution mode from the visual and language context.
Each execution mode is defined by a temporal action quantization level and a prediction horizon, allowing the policy to choose both the granularity of its commands and the temporal extent of its action predictions. 
To validate the effectiveness of \sname{}, we conduct extensive experiments across Robocasa-Kitchen~\citep{nasiriany2024robocasa}, LIBERO~\citep{Liu2023libero}, and DexJoCo~\citep{wang2026dexjoco} simulated robotic benchmarks and six real-world manipulation tasks including long-horizon, dexterous tasks covered on three robot platforms: Franka Research arm with gripper, OpenArm robot arm with two 6-DoF dexterous hands, and humanoid ALLEX robot with two 15-DoF dexterous hands. 
Compared with the baselines, \sname{} achieves up to 1.8$\times$ faster than the base VLA, and 1.75$\times$ throughput on the conveyor package-handling task with the humanoid ALLEX, while maintaining task success rates.

Our contributions are:
\begin{itemize}[leftmargin=*,itemsep=0mm]
\item We highlight the tendency of VLA policies to inherit fine-grained execution patterns from teleoperated demonstrations as a key source of temporal inefficiency.
\item We propose \sname{}, a lightweight architecture that models execution modes with different temporal quantization levels and action horizons using mode-specific action decoders, a quantization judge and a context-conditioned router.
\item We demonstrate consistent gains on simulation benchmarks and real-world tasks, achieving a stronger balance between task success and execution efficiency than the base VLA and fixed temporal quantization baselines.
\end{itemize}

\section{Related Work}
\label{sec:related}

\textbf{Vision-Language-Action models (VLAs).}
VLAs have recently shown great promise in robotic policy learning by leveraging large-scale robot datasets.
One line of VLA models~\citep{kim2024openvla, pertsch2025fast, driess2023palm} discretizes the continuous action space and predicts actions as tokens, showing that pre-trained VLMs~\citep{beyer2024paligemma, chen2026eagle, bai2025qwen3, chen2023pali} can be effectively adapted to low-level action prediction.
In contrast, a growing body of studies~\citep{intelligence2025pi_, li2024cogact} attaches action experts that generate continuous action chunks via diffusion or flow-matching, enabling more accurate predictions.
For example, GR00T N1.5~\citep{nvidia2025gr00t} and RLDX-1~\citep{kim2026rldx} condition a flow-matching head on VLM features to produce action chunks.
In this work, we build on a flow-matching action expert for action generation.
While most diffusion-based VLAs predict action chunks at a single, fixed temporal granularity set by the data collection frequency, this single granularity is not equally suited to all phases of a task.
Motivated by this observation, we explore whether jointly learning actions at  multiple temporal granularities and adapting the action granularity to the current context can lead to faster and more accurate task execution.

\textbf{Faster-than-Demonstration Execution.}
The execution speed of imitation-learned policies is bounded by the pace of their demonstrations, and several lines of work have aimed to exceed it.
Reinforcement-learning-based fine-tuning adjusts the policy using task-level rewards to encourage faster and more decisive behaviors~\citep{yuan2025speedtuning}, but tends to push the policy toward uniformly faster execution.
Another direction modifies the temporal structure of actions, as explored in DemoSpeedup~\citep{guo2025demospeedup} and SAIL~\citep{arachchige2025sail}.
While these approaches improve execution speed, they determine the degree of acceleration either at implicit action entropy signals or through kinematic cues for switching between preset speeds, both derived from action statistics, rather than letting the policy itself decide how much to accelerate from the scene it perceives at inference time.

\textbf{Router-Based Expert Selection.}
Conditional computation employs a lightweight router that selects which sub-network to apply to each input, and has been widely adopted to scale capacity without proportional cost.
In most designs the candidates are structurally identical and differ only in learned weights, whether FFN experts~\citep{fedus2022switch} or attention heads~\citep{jin2024moh}, so a routing decision carries no predefined meaning.
Within VLAs, MoLe-VLA~\citep{zhang2026mole} routes over backbone layers to reduce the cost of each forward pass, leaving the action output unchanged.
Mixture of Horizons~\citep{jing2025mixture} is the closest to our setting, as it constructs action heads at several chunk lengths, but it fuses their outputs through a gate, so each action is still emitted at a single fixed temporal granularity.
None of these, however, treat the temporal granularity of the action as the routing target, where a router selects one decoder per state from a set of decoders that differ in temporal resolution and horizon.

\section{Method}

\begin{figure}[t]
\centering
\includegraphics[width=0.85\linewidth]{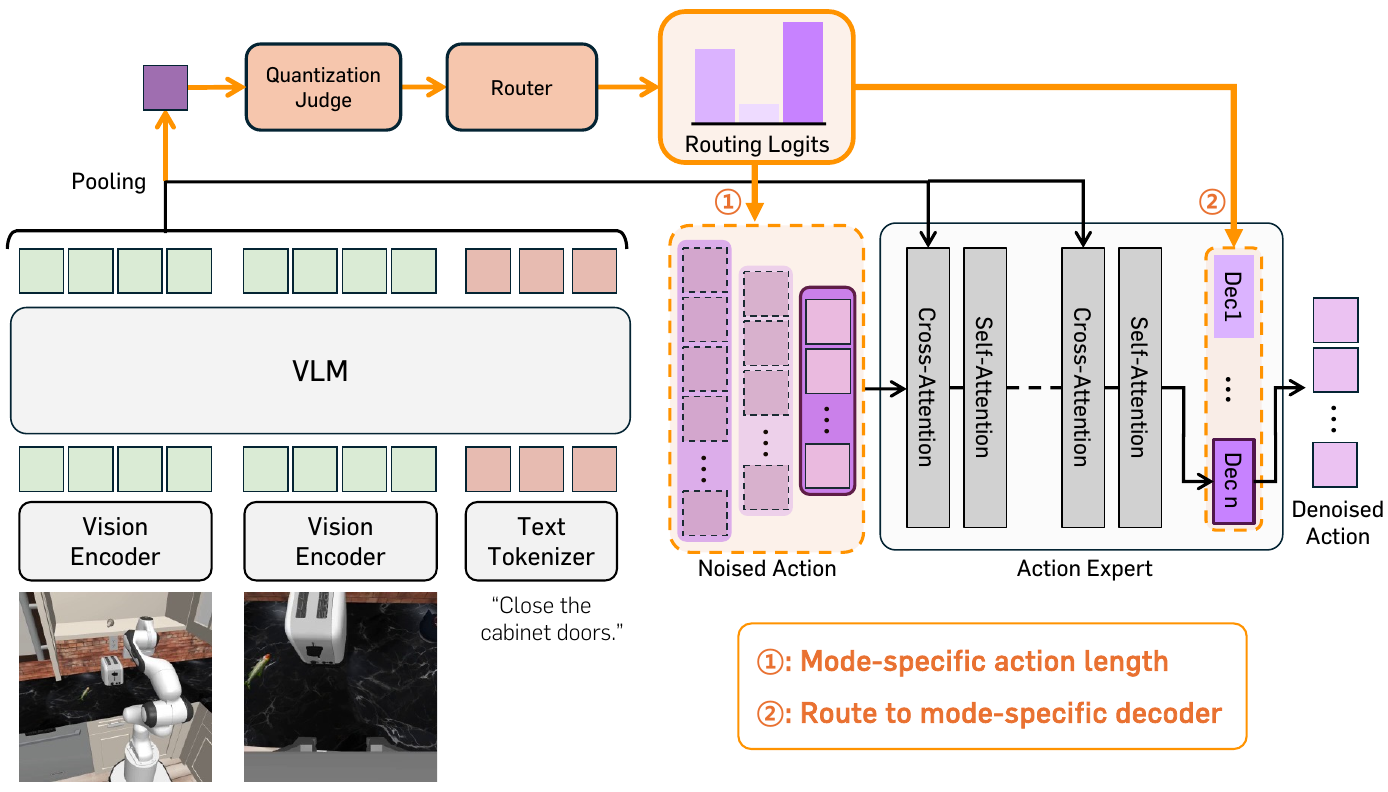}
\caption{\textbf{Overview of \sname{}.} \sname{} is a unified framework for adaptive temporal action execution in pretrained VLAs. It equips a shared action expert with mode-specific action decoders that support different temporal granularities and prediction horizons.}
\label{figure:architecture}
\vspace{-0.2cm}
\end{figure}

In this section, we introduce \sname{}, a unified framework for accelerating robot policy execution through adaptive temporal action quantization. In a nutshell, \sname{} leverages a shared action expert with multiple action decoders that support different temporal granularities and prediction horizons. At each inference step, a learned quantization judge determines whether to use fine-grained or quantized decoding, and a router selects a mode within the chosen group. We also describe the system-level controller configuration used to track the resulting action commands. The overall framework is illustrated in Figure~\ref{figure:architecture}.

Formally, given a language instruction $\mI$, visual observations $\mV_t$, and a proprioceptive robot state $\vs_t$, let $\Phi_t = f_\theta(\mI,\mV_t,\vs_t)$ denote the conditioning context for predicting an action chunk $\mA_t=(\va_{t,1},\ldots,\va_{t,H})$ with action horizon $H$.
We build on a pretrained diffusion-based VLA~\citep{nvidia2025gr00t}, whose action expert uses a diffusion transformer (DiT) backbone~\citep{peebles2023scalable}  trained with flow matching~\citep{lipman2022flow}. Given a clean chunk $\mA_t$, a sampled flow time $\tau$, and Gaussian noise $\epsilon\sim\gN(\vzero,I)$, we construct the noisy chunk $\mA_t^\tau=\tau\mA_t+(1-\tau)\epsilon$. The velocity network $\vu_\theta(\Phi_t,\mA_t^\tau,\tau)$ is trained to predict $\mA_t-\epsilon$:
\begin{equation}
    \gL_{\mathrm{FM}}(\theta)
    =
    \E_{\mA_t,\tau,\epsilon}
    \left[
        \left\|
            \vu_\theta(\Phi_t,\mA_t^\tau,\tau)
            -
            (\mA_t-\epsilon)
        \right\|^2
    \right].
    \label{eq:flow-matching}
\end{equation}
We extend this formulation to multiple temporal action modes using mode-specific output decoders with a shared action expert.
 
\subsection{Temporal Action Quantization}
Temporal action quantization represents a sequence of demonstration actions with fewer action commands by aggregating adjacent actions. We define $K$ action modes. For each mode $k\in\{1,\ldots,K\}$, we convert $H_k$ demonstration steps into $T_k\leq H_k$ commands, with a temporal quantization ratio $q_k=H_k/T_k$.
Here, $H_k$ denotes the source horizon and $T_k$ is the number of predicted commands. Note that \emph{action quantization} refers to aggregating action values along time. 
We construct fine-grained modes with $q_k=1$ and quantized modes with $q_k>1$. Both fine-grained and quantized modes further include shorter and longer source horizons to provide temporal variety.

We partition each source segment into $T_k$ contiguous blocks and aggregate the actions within each block. In detail, action quantization depends on the action representation. For absolute actions, each block retains its final command. For delta-pose actions, we sum translations expressed in a fixed Cartesian frame, compose rotational increments, and retain the final gripper command. 
The resulting target is denoted by $\mA_t^{(k)}\in\mathbb R^{T_k\times d_a}$, where $d_a$ is the action dimension.
Detailed aggregation rules are provided in Appendix~\ref{appendix:temporal_action_quantization_details}.

\subsection{Shared Action Expert with Mode-Specific Action Decoders}
We implement these temporal action modes using a shared flow-matching action expert with mode-specific output decoders. The action encoder and DiT backbone are shared across action modes, while each mode $k$ has a lightweight MLP decoder $D_k$. 
For mode $k$, the shared encoder and DiT backbone process a noisy action sequence of length $T_k$. Decoder $D_k$ maps the resulting features to a flow velocity prediction of shape $T_k \times d_a$, where $d_a$ is the action dimension.
This design allows each mode to learn its corresponding temporal granularity without requiring global temporal resampling of the dataset.
During training, we compute the loss for all $K$ modes using their respective targets and sequence lengths. At inference, action generation uses the sequence length $T_k$ and decoder $D_k$ associated with the selected action mode. 

\subsection{Quantizability-Guided Mode Selection}
We select an action mode using two signals. The quantization judge first explicitly selects whether to use fine-grained or quantized action, and the router then selects a mode within the chosen group.

\paragraph{Group selection and mode routing.}
We partition the modes into a fine-grained group, $\mathcal{K}_{\mathrm{fine}} = \{k : q_k = 1\}$, and a quantized group, $\mathcal{K}_{\mathrm{quant}} = \{k : q_k > 1\}$. The predicted quantizability score determines the action group. A lightweight MLP router produces a set of probabilities $\boldsymbol\alpha_t=(\alpha_{t,1},\ldots,\alpha_{t,K})^\top$ over the modes, and these select the mode within the eligible group.

\paragraph{Quantizability supervision.}  
To explicitly determine whether acceleration is appropriate, we supervise the quantization judge with annotations using task execution outcomes and a large vision-language model (LVLM).
We first examine task behavior under several fixed action quantization settings applied to each task. 
We use these observations as task-level cues to guide the identification of phase-level patterns of quantizability within the demonstrations.
LVLM then refines these patterns into criteria using demonstration samples and applies them to each time step of the original demonstrations, producing a per-step quantizability score $y\in[0,1]$. 
These scores supervise the judge through \cref{eq:conf-loss}, while mode-specific prediction errors supervise the router through \cref{eq:policy-loss}.
The details regarding the creation and actual annotation criteria are provided in the appendix~\ref{appendix:quantizability_details}.

\subsection{Training Objectives}
\label{sec:training}

We jointly optimize action prediction and routing, with an auxiliary regression objective for the quantization judge.
For mode $k$, let $\ell_{k}$ denote the per-sample prediction error in \cref{eq:flow-matching}, evaluated using context $\Phi_t$, aggregated target $\mA_t^{(k)}$, and the velocity network $\vu_\theta^{(k)}$ formed by the shared action expert and decoder $D_k$.
Each mode uses independently sampled noise and flow times.
Prediction errors are masked and averaged over valid action elements, which accounts for the different target action lengths across modes. 

\paragraph{Action policy learning.}
Let $\boldsymbol\alpha=(\alpha_1,\ldots,\alpha_K)^\top\in \mathbb R^K $ denote the router probabilities from current context $\Phi_t$, let $\mathbf w=(w_1,\ldots,w_K)^\top$ denote its training weights, and $\mathbf r=(r_1,\ldots,r_K)^\top \in \mathbb R^K$ a target probability vector. The policy objective is
\begin{equation}
\begin{aligned}
    \gL_{\mathrm{policy}}
    ={}&\mathbb E\!\left[
        \sum_{k=1}^{K}w_k\ell_k
        +\lambda_{\mathrm{sup}}\operatorname{KL}
        (\mathbf r\,\|\,\boldsymbol\alpha)
    \right]
    +\lambda_{\mathrm{bal}}\gL_{\mathrm{bal}}
    \label{eq:policy-loss}
\end{aligned}
\end{equation}
where the target $r_k \propto \exp(-\ell_k/\beta)$ favors modes with lower prediction errors, with temperature $\beta$. The training weights $\mathbf w$ mix $\boldsymbol\alpha$ with a uniform distribution, and $\gL_{\mathrm{bal}}$ penalizes their deviation from uniform, preventing collapse onto a single decoder~\citep{fedus2022switch}.

\paragraph{Quantizability regression.}
The judge $J_\psi$ regresses the LVLM-generated quantizability score $y$ at the first frame of each training chunk:
\begin{equation}
    \gL_{\mathrm{conf}}
    =\mathbb E\!\left[
        \left(J_\psi(\operatorname{sg}(\Phi_t))-y\right)^2
    \right],
    \label{eq:conf-loss}
\end{equation}
where $\operatorname{sg}$ denotes stop-gradient, so that $\gL_{\mathrm{conf}}$ updates only the judge parameters $\psi$. At inference, group selection uses the judge's predicted score.

The overall training objective is
\begin{equation}
    \gL=\gL_{\mathrm{policy}}+\lambda_{\mathrm{conf}}\gL_{\mathrm{conf}}.
    \label{eq:total-loss}
\end{equation}

\subsection{Controller Configuration for Accelerated Execution}

Temporally quantized actions require controllers to track larger commanded changes. In our experiment, insufficient tracking leads to execution failures. We adjust the tracking and gripper gains to support accelerated execution. Controller-specific settings are reported in Appendix~\ref{appendix:controller_config_sim}.

\section{Experiments}
\label{sec:experiments}

We validate the performance of \sname and the effect of the proposed components through extensive experiments. In particular, we design our experiments to investigate the following questions: 

\begin{itemize}[leftmargin=*,itemsep=0mm]
\item Does \sname{} improve the trade-off between task success and execution efficiency on simulation benchmarks with various action representations? (Table~\ref{tab:sim_libero_robocasa}, ~\ref{tab:sim_dexjoco})

\item Can \sname accelerate real-world dexterous and long-horizon manipulation while maintaining task success across different embodiments?
(Table~\ref{tab:real_world_droid_openarm}, ~\ref{tab:real_world_allex})

\item How does each component of \sname{} contribute to performance? (Table ~\ref{tab:ablation})
\end{itemize}

\subsection{Experimental Setups}

\paragraph{Datasets.}
For simulation, we use RoboCasa-Kitchen~\citep{nasiriany2024robocasa}, LIBERO~\citep{Liu2023libero}, and DexJoCo~\citep{wang2026dexjoco}.
We evaluate 24 RoboCasa-Kitchen tasks using 300 demonstrations per task, and report the results separately for pick-and-place tasks and the remaining tasks.
For LIBERO, we evaluate the Spatial, Object, Goal, and Long suites.
For DexJoCo, we evaluate a subset of six single-arm tasks that require contact-rich dexterous manipulation.
In our setup, RoboCasa-Kitchen and LIBERO use delta actions, whereas DexJoCo uses absolute actions, allowing us to assess temporal action quantization under both representations.
For real-world evaluation, we evaluate six tasks across three embodiments.
On Franka arm with a gripper, we evaluate \textit{Pick-and-Place}, \textit{Air Fryer}, and \textit{Stack Cup}.
\textit{Air Fryer} is a long-horizon task that requires opening the basket, placing food inside, and closing it.
\textit{Stack Cup} requires two consecutive stacking operations.
On a bimanual OpenArm equipped with 6-DoF dexterous hands, we evaluate two bimanual tasks, \textit{Table Sweeping} and \textit{Pump-and-Scrub}.
\textit{Table Sweeping} requires sweeping a table using a dustpan and brush, and
\textit{Pump-and-Scrub} involves dispensing soap and wiping a plate.
On the ALLEX humanoid equipped with 15-DoF dexterous hands, we evaluate long-horizon \textit{Conveyor Package Handling}, in which the robot handles 10 packages in a row per trial, re-orienting each arriving package so that its shipping label faces forward.
Detailed task descriptions, datasets, and robot configurations are provided in Appendices~\ref{appendix:task_details} and~\ref{appendix:hw_details}.

\begin{figure}[t]
\centering
\includegraphics[width=0.99\linewidth]{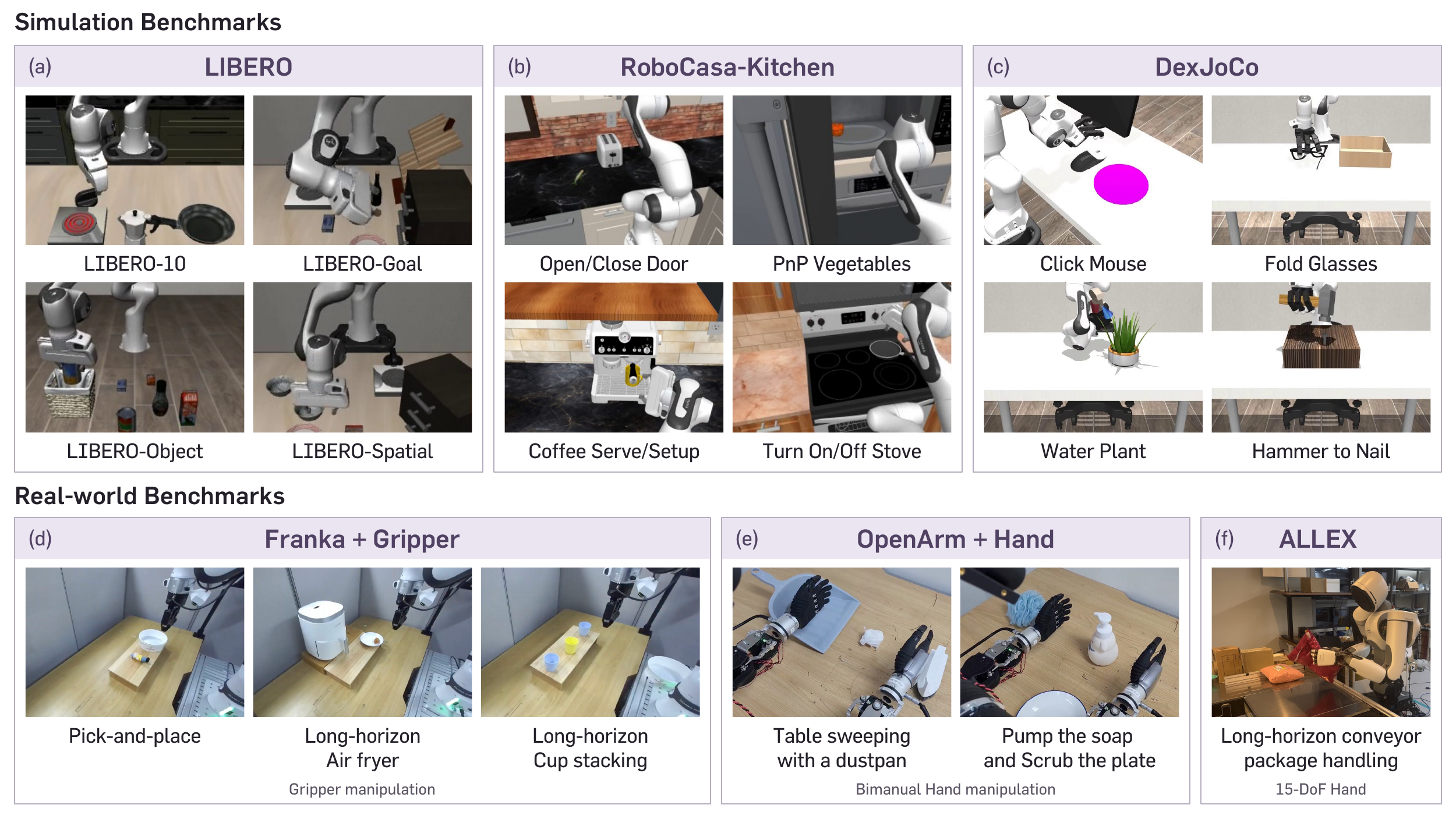}
\vspace{-0.2cm}
\caption{\textbf{Experimental setup.}
Top: Three simulation benchmarks, LIBERO, RoboCasa-Kitchen, and DexJoCo.
Bottom: Real-world manipulation tasks across three embodiments,
a Franka Research Arm with a gripper, an OpenArm robot with a
dexterous hand, and an ALLEX humanoid.
The real-world evaluation includes long-horizon and dexterous
manipulation tasks.}
\vspace{-0.2cm}
\end{figure}

\paragraph{Implementation Details.}
We implement \sname{} by fine-tuning pre-trained Vision-Language-Action models (VLA). Specifically, we use GR00T N1.5~\citep{nvidia2025gr00t} and RLDX-1-IMG~\citep{kim2026rldx} by following their official implementation. We use four action decoders, corresponding to two fine-grained and two quantized modes, each with a longer and a shorter source horizon, and the quantization judge is supervised using Large-Vision-Language-Model (LVLM)-generated annotations.
Training hyperparameters, action mode configurations, annotation procedures, and controller settings are provided in Appendix~\ref{appendix:exp_details}.

\paragraph{Baselines.}
We compare the original model with variants incorporating three acceleration methods and \sname{}.
Action Quantization~\citep{tang2025vlash} applies fixed temporal action quantization with a ratio of 2 over the entire trajectory on inference. 
DemoSpeedup~\citep{guo2025demospeedup} accelerates demonstrations through entropy-guided temporal downsampling, and SAIL~\citep{arachchige2025sail} combines action scheduling, temporally consistent prediction, and speed modulation to support faster policy execution.
We use GR00T N1.5~\citep{nvidia2025gr00t} as the backbone for all experiments, and additionally use RLDX-1-IMG~\citep{kim2026rldx} on the simulation benchmarks to verify that \sname{} generalizes across VLAs. Baseline adaptations are detailed in Appendix ~\ref{appendix:adaptation_sail} and evaluation protocols are detailed in Appendix ~\ref{appendix:task_details} and ~\ref{appendix:sim_hyperparams_details}.

\paragraph{Metrics.}
We report the success rate (SR, \%) for every task, with partial score on Stack Cup and OpenArm tasks as detailed in Appendix~\ref{appendix:exp_details}.
In simulation, we report throughput-with-regret (TPR)~\citep{arachchige2025sail}, which scores each trial by how quickly it succeeds and assigns every failure the same penalty, together with an effective speedup (ES).
Following the mean picks per hour (MPPH) metric of Dex-Net~\citep{mahler2019learning}, we define ES as the ratio of a method's success throughput to that of its base model, where each failed trial is charged the full episode budget $t^{\max}$.
A method that moves faster but fails more often can therefore have ES below 1.
On the Franka and OpenArm tasks, we report the mean wall-clock execution time (Time, s) over fully successful trials, and on the ALLEX conveyor task, the number of packages handled successfully per minute (Thpt.).
Full definitions are given in Appendix~\ref{appendix:metrics}.

\subsection{Main Results}

\paragraph{Simulation Benchmarks.}

\begin{table}[t!]
\centering
\captionsetup{skip=2pt}
\setlength{\tabcolsep}{2.5pt}
\renewcommand{\arraystretch}{1.1}
\caption{\textbf{Simulation benchmark results.}
    We compare \sname{} with baseline methods on RoboCasa-Kitchen and LIBERO.
    For RoboCasa-Kitchen, we report results on the pick-and-place tasks (PnP) and the
    remaining tasks (Others), with models trained using 300 demonstrations per task.
    For LIBERO, we report results on the short-horizon suites (Short, averaged over Spatial,
    Object, and Goal) and the long-horizon suite (Long). We report the success rate (SR, \%),
    throughput-with-regret (TPR, $\times 10^{-3}$), and the
    effective speedup (ES).
    Base models are listed for reference ($\mathrm{ES}=1$ by definition).
    Among the acceleration methods for each base model, the best and second-best results are
    in \textbf{bold} and \underline{underlined}, respectively.}
\label{tab:sim_libero_robocasa}
\resizebox{\textwidth}{!}{
\begin{tabular}{l ccc ccc ccc ccc}
    \toprule
    \multirow{3}{*}{Method}
    & \multicolumn{6}{c}{\textbf{RoboCasa-Kitchen}} & \multicolumn{6}{c}{\textbf{LIBERO}} \\
    \cmidrule(lr){2-7} \cmidrule(lr){8-13}
    & \multicolumn{3}{c}{PnP} & \multicolumn{3}{c}{Others}
    & \multicolumn{3}{c}{Short} & \multicolumn{3}{c}{Long} \\
    \cmidrule(lr){2-4} \cmidrule(lr){5-7} \cmidrule(lr){8-10} \cmidrule(lr){11-13}
    & SR $\uparrow$ & TPR $\uparrow$ & ES $\uparrow$
    & SR $\uparrow$ & TPR $\uparrow$ & ES $\uparrow$
    & SR $\uparrow$ & TPR $\uparrow$ & ES $\uparrow$
    & SR $\uparrow$ & TPR $\uparrow$ & ES $\uparrow$ \\
    \midrule
    RLDX-1-IMG\,[\citenum{kim2026rldx}]
        & 48.5 & 1.10 & 1.00 & 76.3 & 2.73 & 1.00 & 96.7 & 7.93 & 1.00 & 88.6 & 3.15 & 1.00 \\
    \midrule
    \quad + Fixed Quantization\,[\citenum{tang2025vlash}]
        & \underline{43.0} & \textbf{1.55} & \underline{0.87}
        & \underline{72.0} & \textbf{4.00} & \underline{0.97}
        & 89.4 & \underline{11.80} & 1.23
        & \underline{79.2} & \underline{4.44} & \underline{1.10} \\
    \quad + DemoSpeedUp\,[\citenum{guo2025demospeedup}]
        & 28.7 & 0.80 & 0.49
        & 62.7 & \underline{3.37} & 0.68
        & 93.5 & \textbf{14.03} & \textbf{1.54}
        & 75.2 & \textbf{4.95} & 1.06 \\
    \quad + SAIL\,[\citenum{arachchige2025sail}]
        & 38.5 & 1.03 & 0.72
        & 66.6 & 2.80 & 0.75
        & \underline{96.5} & 10.34 & 1.27
        & 76.6 & 3.38 & 0.92 \\
    \rowcolor{violet!10}
    \quad \textbf{+ \sname{}}
        & \textbf{45.0} & \underline{1.54} & \textbf{0.93}
        & \textbf{76.9} & 3.21 & \textbf{1.07}
        & \textbf{97.7} & 10.36 & \underline{1.29}
        & \textbf{90.2} & \underline{4.44} & \textbf{1.31} \\
    \midrule
    GR00T N1.5\,[\citenum{nvidia2025gr00t}]
        & 52.3 & 1.05 & 1.00 & 72.4 & 2.20 & 1.00 & 97.8 & 8.36 & 1.00 & 89.2 & 3.12 & 1.00 \\
    \midrule
    \quad + Fixed Quantization\,[\citenum{tang2025vlash}]
        & \underline{49.0} & \underline{1.56} & \underline{0.96}
        & \underline{63.1} & \underline{2.92} & \underline{0.81}
        & 91.2 & \textbf{12.41} & \underline{1.25}
        & \underline{82.0} & \textbf{4.78} & \underline{1.21} \\
    \quad + DemoSpeedUp\,[\citenum{guo2025demospeedup}]
        & 40.3 & 1.15 & 0.71
        & 57.3 & 2.77 & 0.67
        & 87.7 & \underline{12.09} & 1.14
        & 71.6 & 3.91 & 0.90 \\
    \quad + SAIL\,[\citenum{arachchige2025sail}]
        & 32.5 & 0.77 & 0.52
        & 58.5 & 2.31 & 0.68
        & \underline{95.5} & 10.85 & 1.23
        & 79.0 & 3.89 & 1.03 \\
    \rowcolor{violet!10}
    \quad \textbf{+ \sname{}}
        & \textbf{55.0} & \textbf{1.64} & \textbf{1.16}
        & \textbf{72.9} & \textbf{3.30} & \textbf{1.14}
        & \textbf{97.1} & 10.95 & \textbf{1.27}
        & \textbf{89.2} & \underline{4.27} & \textbf{1.26} \\
    \bottomrule
\end{tabular}
}
\vspace{-0.6em}
\end{table}

Tables~\ref{tab:sim_libero_robocasa} and~\ref{tab:sim_dexjoco} compare \sname{} with the acceleration baselines across RoboCasa-Kitchen, LIBERO, and DexJoCo.
Across both VLA backbones, \sname{} largely preserves the success rate of the base policy while improving execution efficiency, achieving an effective speedup above $1\times$ in all but one setting.
In contrast, the baseline acceleration methods often attain high raw throughput by sacrificing task success, resulting in an effective speedup below $1\times$ in several settings.
This trend is consistent on DexJoCo, where \sname{} improves both success rate and effective speedup.
Overall, \sname{} provides a more favorable balance between task success and execution efficiency across action representations and VLA backbones.

\begin{table}[!t]
\centering
\captionsetup{skip=2pt}
\setlength{\tabcolsep}{3pt}
\renewcommand{\arraystretch}{1.1}
\caption{\textbf{Real-World evaluation results.}
    We report average success rate (SR, \%, over 32 trials per task; partial scores of 0.5 are given on Stack Cup, Table Sweeping, and Pump and Scrub, see Appendix~\ref{appendix:exp_details})
    and execution time (Time, average wall-clock execution time on fully successful trials)
    on five real-world tasks across two embodiments: three Franka+Gripper tasks
    (\textit{Pick-and-Place, Air Fryer, Stack Cup}) and two OpenArm+Hand tasks
    (\textit{Table Sweeping, Pump and Scrub}). 
}\label{tab:real_world_droid_openarm}
\resizebox{\textwidth}{!}{
\begin{tabular}{lcccccccccc}
    \toprule
    & \multicolumn{6}{c}{\textbf{Franka + Gripper}} & \multicolumn{4}{c}{\textbf{OpenArm + Bimanual Hand}} \\
    \cmidrule(lr){2-7} \cmidrule(lr){8-11}
    & \multicolumn{2}{c}{Pick-and-Place} & \multicolumn{2}{c}{Air Fryer} & \multicolumn{2}{c}{Stack Cup}
    & \multicolumn{2}{c}{Table Sweeping} & \multicolumn{2}{c}{Pump and Scrub} \\
    \cmidrule(lr){2-3} \cmidrule(lr){4-5} \cmidrule(lr){6-7} \cmidrule(lr){8-9} \cmidrule(lr){10-11}
    Method
    & SR $\uparrow$ & Time (s) $\downarrow$
    & SR $\uparrow$ & Time (s) $\downarrow$
    & SR $\uparrow$ & Time (s) $\downarrow$
    & SR $\uparrow$ & Time (s) $\downarrow$
    & SR $\uparrow$ & Time (s) $\downarrow$ \\
    \midrule
    GR00T N1.5\,[\citenum{nvidia2025gr00t}] & 81.3 & 12.1 & 81.3 & 24.6 & 84.4 & 35.3 & 62.5 & 21.9 & 64.1 & 30.0 \\
    \midrule
    \quad + Fixed Quantization\,[\citenum{tang2025vlash}] & 62.5 & \underline{7.7} & 53.1 & \textbf{14.2} & \underline{62.5} & \textbf{18.7} & 48.4 & \textbf{13.3} & 42.2 & 19.9 \\
    \quad + DemoSpeedUp\,[\citenum{guo2025demospeedup}] & \underline{68.8} & 8.1 & \underline{71.9} & 18.9 & 53.1 & 24.2 & \underline{56.3} & 17.9 & \underline{53.1} & 20.2 \\
    \quad + SAIL\,[\citenum{arachchige2025sail}] & 65.6 & 7.9 & 56.3 & 16.8 & 45.3 & 20.0 & 46.9 & 15.4 & 45.3 & \textbf{15.3} \\
    \rowcolor{violet!10}\quad + \textbf{\sname} & \textbf{84.4} & \textbf{7.6} & \textbf{81.3} & \underline{14.6} & \textbf{81.3} & \underline{19.4} & \textbf{59.4} & \underline{14.6} & \textbf{65.6} & \underline{18.2} \\
    \bottomrule
\end{tabular}
}
\vspace{-0.3em}
\end{table}

\paragraph{Real-World Tasks.}
Tables~\ref{tab:real_world_droid_openarm} and~\ref{tab:real_world_allex} compare \sname{} with the baselines across the three real-world embodiments.
On the Franka tasks, \sname{} reduces execution time by 37--45\% while preserving a success rate comparable to or higher than that of the base VLA.
Notably, on the Air Fryer task, it reduces execution time from 24.6\,s to 14.6\,s without degrading success.
Fixed Action Quantization achieves comparable or slightly shorter execution on some tasks, but drops substantial success rate by 28.2 and 21.9 percentage points on Air Fryer and Stack Cup, respectively.
On the bimanual OpenArm tasks, \sname{} reduces execution time by 33--39\% while keeping the base model's success rate.
Finally, on the ALLEX conveyor task, \sname{} increases throughput from 3.6 to 6.3 packages per minute while improving success rate from 86\% to 94\%.
These results show that \sname{} transfers across gripper-based, bimanual, and humanoid manipulation without the substantial success rate degradation observed with the baseline acceleration methods.

\begin{table}[t!]
\centering
\renewcommand{\arraystretch}{1.1}
\begin{minipage}[b]{0.49\textwidth}
\centering
\setlength{\tabcolsep}{4pt}
\caption{\textbf{DexJoCo benchmark results.}
    We compare \sname{} with baselines on DexJoCo, where a dexterous hand performs
    contact-rich manipulation tasks. We report the success rate (SR, \%),
    throughput-with-regret (TPR, $\times 10^{-3}$), and the effective speedup (ES) over 6 single-arm tasks.
    Among the acceleration methods, the best and second-best results are in \textbf{bold} and \underline{underlined}, respectively.}
\vspace{-0.1cm}
\label{tab:sim_dexjoco}
\small
\begin{tabular}[b]{lccc}
    \toprule
    \multirow{2}{*}{Method}
    & \multicolumn{3}{c}{\textbf{DexJoCo (Single-arm)}} \\
    \cmidrule(lr){2-4}
    & SR $\uparrow$ & TPR $\uparrow$ & ES $\uparrow$ \\
    \midrule
    GR00T N1.5\,[\citenum{nvidia2025gr00t}]             & 76.3 & 1.83 & 1.00 \\
    \midrule
    \quad + Fixed Quantization\,[\citenum{tang2025vlash}]                         & 63.0 & 1.65 & 0.70 \\
    \quad + DemoSpeedUp\,[\citenum{guo2025demospeedup}] & \underline{65.7} & \underline{1.75} & \underline{0.76} \\
    \quad + SAIL\,[\citenum{arachchige2025sail}]        & 60.3 & 1.58 & 0.65 \\
    \rowcolor{violet!10}
    \quad \textbf{+ \sname{}}                           & \textbf{77.3} & \textbf{2.22} & \textbf{1.11} \\
    \bottomrule
\end{tabular}
\end{minipage}\hfill
\begin{minipage}[b]{0.47\textwidth}
\centering
\setlength{\tabcolsep}{5pt}
\caption{\textbf{Real-world evaluation results.}
We compare \sname{} with baselines on Conveyor-based package handling task with a humanoid ALLEX (15-DoF hand). We report the success rate (SR, \%) and throughput (Thpt., packages/min). Among the acceleration methods, the best and second-best results are in \textbf{bold} and \underline{underlined}, respectively.} 
\label{tab:real_world_allex}
\small
\begin{tabular}[b]{lcc}
    \toprule
    \multirow{2}{*}{Method}
    & \multicolumn{2}{c}{\makecell{\textbf{Conveyor}\\\textbf{Package Handling}}} \\
    \cmidrule(lr){2-3}
    & SR $\uparrow$ & Thpt. $\uparrow$ \\
    \midrule
    GR00T N1.5\,[\citenum{nvidia2025gr00t}]             & 86.0 & 3.6 \\
    \midrule
    \quad + Fixed Quantization\,[\citenum{tang2025vlash}]                         & 72.0 & 4.2 \\
    \quad + DemoSpeedUp\,[\citenum{guo2025demospeedup}] & \underline{86.0} & \underline{4.7} \\
    \quad + SAIL\,[\citenum{arachchige2025sail}]        & 78.0 & 4.3 \\
    \rowcolor{violet!10}
    \quad \textbf{+ \sname{}}                           & \textbf{94.0} & \textbf{6.3} \\
    \bottomrule
\end{tabular}
\end{minipage}
\vspace{-0.4cm}
\end{table}

\subsection{More Analysis}

\begin{table*}[t]
\centering
\captionsetup{skip=2pt}
\caption{
\textbf{Ablation study.}
Success rate (SR, \%), throughput-with-regret (TPR, $\times 10^{-3}$), and effective speedup (ES) on RoboCasa-Kitchen, averaged over all 24 tasks. ES is computed relative to the base model GR00T N1.5. Our default setting is highlighted in \colorbox{violet!10}{violet}.}
\label{tab:ablation}

\phantomsubcaption\label{tab:ablation:a}
\phantomsubcaption\label{tab:ablation:b}
\phantomsubcaption\label{tab:ablation:c}

\renewcommand{\arraystretch}{1.05}

\resizebox{\textwidth}{!}{%
\begin{tabular}[t]{@{}c@{}}
    \setlength{\tabcolsep}{4pt}%
    \begin{tabular}{lccc}
        \toprule
        Method & SR $\uparrow$ & TPR $\uparrow$ & ES $\uparrow$ \\
        \midrule
        w/o Mode-specific Decoders & 63.3 & 2.44 & 1.00 \\
        w/o Quantization Judge     & 65.1 & 2.01 & 1.01 \\
        \rowcolor{violet!10}
        \sname{} & \textbf{66.9} & \textbf{2.75} & \textbf{1.14} \\
        \bottomrule
    \end{tabular} \\
    \noalign{\vskip 0.5em}
    (a) Component ablation.
\end{tabular}%
\hspace{1em}%
\begin{tabular}[t]{@{}c@{}}
    \setlength{\tabcolsep}{4pt}%
    \begin{tabular}{lccc}
        \toprule
        Method & SR $\uparrow$ & TPR $\uparrow$ & ES $\uparrow$ \\
        \midrule
        w/o KL      & 63.7 & 2.04 & 0.98 \\
        w/o Balance & 64.8 & 2.38 & 1.04 \\
        \rowcolor{violet!10}
        Balance + KL & \textbf{66.9} & \textbf{2.75} & \textbf{1.14} \\
        \bottomrule
    \end{tabular} \\
    \noalign{\vskip 0.5em}
    (b) Loss design.
\end{tabular}%
\hspace{1em}%
\begin{tabular}[t]{@{}c@{}}
    \setlength{\tabcolsep}{4pt}%
    \begin{tabular}{lccc}
        \toprule
        Method & SR $\uparrow$ & TPR $\uparrow$ & ES $\uparrow$ \\
        \midrule
        Attn. Pool & 64.6 & 2.38 & 1.04 \\
        Last Token & 64.2 & 2.43 & 1.03 \\
        \rowcolor{violet!10}
        Avg. Pool & \textbf{66.9} & \textbf{2.75} & \textbf{1.14} \\
        \bottomrule
    \end{tabular} \\
    \noalign{\vskip 0.5em}
    (c) Router input.
\end{tabular}%
}

\vspace{-0.4cm}
\end{table*}

\paragraph{Ablation Study.}
First, we remove each of the two key components one at a time.
(i) Without the mode-specific decoders, we use the base VLA's single decoder and directly quantize its predicted chunk with the same aggregation rule when the judge selects coarse execution.
(ii) Without the quantization judge, the router selects among all modes.
As shown in Table~\ref{tab:ablation:a}, removing the mode-specific decoders reduces the success rate from 66.9\% to 63.3\% and the effective speedup from 1.14 to 1.00, indicating that dedicated decoders are important for jointly preserving task success and execution efficiency under temporal quantization.
This is consistent with Appendix~\ref{appendix:quantitative_gt_analysis}, where dedicated decoders predict quantized actions more accurately than directly quantizing the base model's predictions.
Removing the quantization judge lowers TPR from 2.75 to 2.01 and ES from 1.14 to 1.01, showing that explicit quantizability supervision is necessary to decide when coarse execution is appropriate.
Next, we examine the loss design in Table~\ref{tab:ablation:b}.
Removing the KL objective causes the largest degradation, reducing SR, TPR, and ES to 63.7\%, 2.04, and 0.98, respectively. Removing the balancing objective also degrades all three metrics, showing that both objectives contribute to learning effective mode selection.
Lastly, Table~\ref{tab:ablation:c} compares router input representations.
Average pooling achieves the highest SR, TPR, and ES, outperforming both attention pooling and the last-token representation, and we therefore use it as the default router input.

\begin{figure}[t]
    \centering
    \includegraphics[width=0.95\linewidth]{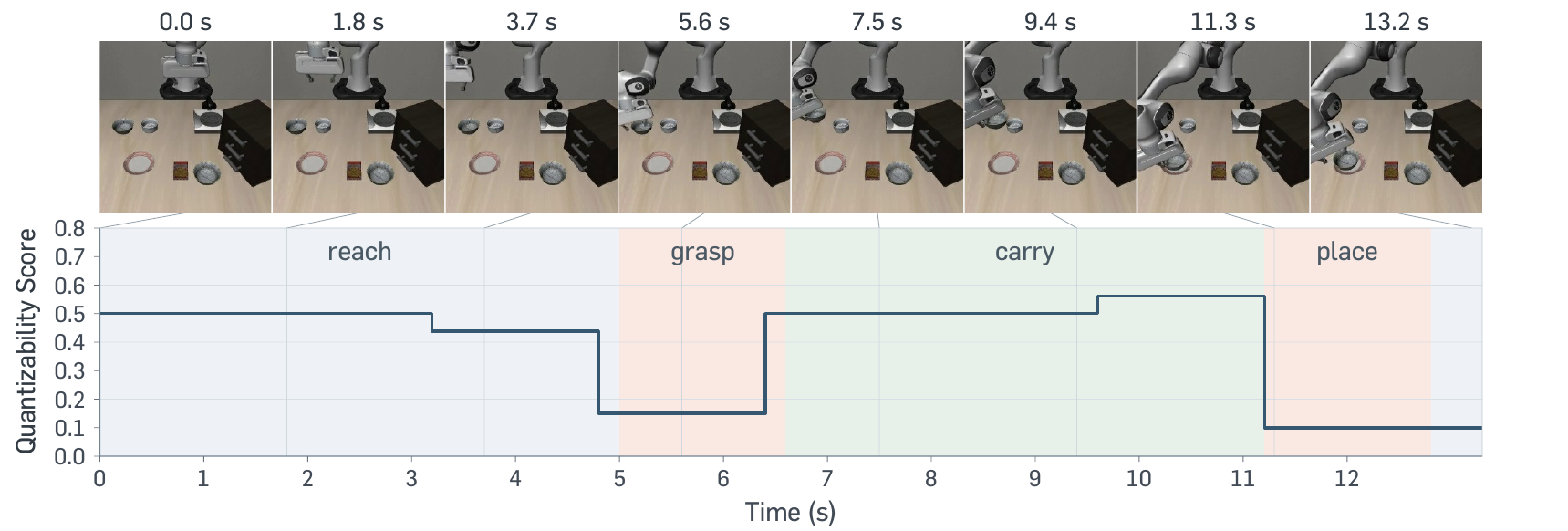}
    \vspace{-.5em}
    \caption{\textbf{Quantizability supervision analysis.}
        We visualize the quantizability scores annotated on a RoboCasa-Kitchen demonstration. Shaded bands indicate manipulation phases.
        Scores are low around contact transitions (grasp and place) and higher during reach and carry, so coarse execution is suppressed exactly where precise control matters.
    }

    \label{fig:analysis}
    \vspace{-0.3cm}
\end{figure}

\paragraph{Qualitative Analysis.}
Figure~\ref{fig:analysis} visualizes the quantizability scores assigned along a representative RoboCasa-Kitchen demonstration.
The quantizability scores remain higher during reaching and carrying, where coarse execution is generally tolerable, and decrease around grasping and placement, where finer control is required. The score also drops during the final approach to the object, indicating that the annotation captures changes in temporal precision requirements across phases.

\paragraph{Limitation.}
\begin{wraptable}{r}{0.42\textwidth}
\vspace{-1.2em}
\centering
\caption{\textbf{Inference cost.} Per-step latency on RoboCasa-Kitchen.}
\label{tab:inference-cost}
\footnotesize
\renewcommand{\arraystretch}{1.05}
\setlength{\tabcolsep}{5pt}
\begin{tabular}{lcc}
    \toprule
    Method & Latency (ms) $\downarrow$ & Rel. Cost \\
    \midrule
    GR00T N1.5 & 40.07 & $1.00\times$ \\
    \rowcolor{violet!10}
    \sname{} & 42.15 & $1.05\times$ \\
    \bottomrule
\end{tabular}
\vspace{-1em}
\end{wraptable}

Table~\ref{tab:inference-cost} compares the per-step inference latency, averaged over 100 runs on an RTX 5090 GPU.
\sname{} incurs a slight overhead over the base model ($1.05\times$), as the judge and router must run before action generation.
However, this $\sim$2\,ms cost is negligible compared to the execution time saved (\eg, 24.6\,s $\rightarrow$ 14.6\,s on Air Fryer), so the efficiency gain far outweighs the added inference cost.

\section{Conclusion}

In this work, we have presented \sname{}, a unified framework for adaptive temporal action execution in Vision-Language-Action models (VLAs).
Motivated by the observation that VLAs inherit the fine-grained execution patterns of teleoperated demonstrations, and that fixed temporal action quantization trades task success for speed, we treat the temporal granularity and prediction horizon of actions as a policy decision rather than a fixed design choice.
\sname{} equips a shared action expert with mode-specific decoders operating at different temporal granularities and prediction horizons, and selects among them at each inference step through a quantizability-guided judge and a context-conditioned router.
Our experiments on three simulation benchmarks and six real-world tasks across gripper-based, dexterous-hand, and humanoid platforms show that \sname{} substantially shortens execution while maintaining the task success of the base VLA, unlike fixed or rule-based acceleration methods, with only marginal inference overhead.
We hope that our work facilitates future research toward generalist robot policies that adapt their execution pace to the task at hand, enabling faster and more reliable deployment in practice.

\clearpage
\section*{AI use statement}

In this work, we used generative AI tools as part of the quantizability supervision pipeline. Claude Opus 5~\citep{anthropic2026opus5} was used to help develop and refine the quantizability criteria and annotation prompts based on task-level evaluation outcomes and demonstration examples. Gemini 3.8 Flash~\citep{googledeepmind2026gemini38flash} was subsequently used through its API to annotate demonstration chunks according to these criteria. The resulting ratings were aggregated into quantizability scores used to supervise the quantization judge, as described in Appendix~\ref{appendix:quantizability_details}. We also used generative AI tools for grammar checking and minor readability edits of author-written text. The authors reviewed the AI-assisted research process and resulting manuscript content and take responsibility for the final content of this work.

\section*{Reproducibility Statement}
We provide hyperparameter details in Section~\ref{sec:experiments} and Appendix~\ref{appendix:sim_hyperparams_details}. In addition, we provide prompts in Appendix~\ref{appendix:prompts}.

\bibliographystyle{plainnat}
\bibliography{references}

\clearpage
\appendix
\setlength{\parskip}{0.5\baselineskip}
\setlength{\textfloatsep}{\origtextfloatsep}
\setlength{\floatsep}{\origfloatsep}
\setlength{\intextsep}{\origintextsep}
\titlespacing*{\section}{0pc}{3ex plus4pt minus3pt}{5pt}
\titlespacing*{\subsection}{0pc}{2.5ex plus3pt minus2pt}{2pt}
\clearpage
\section{Temporal Action Quantization Details}
\label{appendix:temporal_action_quantization_details}

The aggregation rule depends on the action representation.
For a source segment of $H$ actions quantized into $T$ commands, let
$0=s_0<s_1<\cdots<s_T=H$ denote the boundaries of the $T$ contiguous temporal blocks.
The $i$-th block therefore contains
$\{\mathbf{a}_{s_{i-1}+1},\ldots,\mathbf{a}_{s_i}\}$.
When $H$ is not divisible by $T$, we partition the source segment into $T$ contiguous blocks whose sizes differ by at most one.
For absolute actions, each block is represented by its final command:
\begin{equation}
    \bar{\mathbf a}_i = \mathbf a_{s_i}.
    \label{eq:absolute_quantization}
\end{equation}
This preserves the endpoint while removing intermediate commands.

For delta-pose actions, directly summing all action dimensions is generally
invalid because rotational increments do not compose additively.
We write each action as
$\mathbf a_j=(\Delta\mathbf p_j,\mathbf u_j,g_j)$,
where $\Delta\mathbf p_j$ is the translational increment,
$\mathbf u_j$ is the rotational command, and $g_j$ is the gripper command.
Translations expressed in a fixed Cartesian frame can be summed directly,
whereas rotations must be composed on $\mathrm{SO}(3)$.

Let $\phi$ denote the mapping from the policy's rotational command to the rotation vector used by the controller, and define
$\mathbf R_j=\operatorname{Exp}(\phi(\mathbf u_j))$.
Here, $\mathrm{Exp}:\mathbb{R}^3\rightarrow\mathrm{SO}(3)$ and
$\mathrm{Log}:\mathrm{SO}(3)\rightarrow\mathbb{R}^3$ denote the mappings between rotation vectors and rotation matrices.
Under the spatial-increment convention, the quantized action for block $i$ is
\begin{equation}
    \bar{\mathbf a}_i =
    \left(
        \sum_{j=s_{i-1}+1}^{s_i}\Delta\mathbf p_j,\;
        \phi^{-1}\!\left[
            \operatorname{Log}\!\left(
                \mathbf R_{s_i}\cdots\mathbf R_{s_{i-1}+1}
            \right)
        \right],\;
        g_{s_i}
    \right).
    \label{eq:delta_quantization}
\end{equation}

Because rotations are noncommutative, directly adding rotational increments
does not in general reproduce their ordered composition.
The SO(3) composition in \cref{eq:delta_quantization} preserves the net commanded rotation of the original block under the controller convention, although quantization may alter the intermediate trajectory.

When actions are normalized for policy learning, we first transform them back to controller coordinates, perform the aggregation above, and then normalize the resulting actions again.
The resulting quantized target chunk is denoted by
$\mathbf A_t^{(k)}\in\mathbb{R}^{T_k\times d_a}$.

\section{Evaluation Metrics}
\label{appendix:metrics}
In this section, we describe the two efficiency metrics used in the simulation benchmarks in detail. 

\paragraph{Throughput-with-regret (TPR).}
Following SAIL~\citep{arachchige2025sail}, TPR is defined as
\begin{equation}
    \mathrm{TPR} = \frac{1}{N}\sum_{i=1}^{N}
    \left( \frac{S_i}{t_i} - \frac{1-S_i}{t^{\max}} \right),
    \label{eq:tpr}
\end{equation}
where $N$ is the number of trials, $S_i \in \{0,1\}$ indicates whether trial $i$ succeeds, $t_i$ is its number of executed action steps, and $t^{\max}$ is the maximum number of steps allowed per trial, after which the trial is terminated and counted as a failure. TPR rewards faster successes while penalizing all failures equally.

\paragraph{Effective speedup (ES).}
Since TPR penalizes a failure by only $1/t^{\max}$, it underweights the time lost to failed trials. We therefore additionally report
\begin{equation}
    \mathrm{ES} = \frac{T_{\mathrm{base}}}{T}, \qquad
    T = L + \frac{1-\mathrm{SR}}{\mathrm{SR}} \cdot t^{\max},
    \label{eq:es}
\end{equation}
where $\mathrm{SR}\in(0,1]$ denotes the success rate expressed as a fraction, $L$ is the mean number of action steps over successful trials, $T$ is the expected number of action steps per success including the $t^{\max}$ steps consumed by each failed trial, and $T_{\mathrm{base}}$ is $T$ of the corresponding base model. 
Equivalently, $1/T$ is the number of successes per action step, i.e., the success
throughput, following the mean-picks-per-hour (MPPH) metric of Dex-Net~\citep{mahler2019learning}.
$\mathrm{ES}>1$ indicates that a method completes tasks faster than its base model once failures are taken into account, and $\mathrm{ES}=1$ for the base model by definition.

\paragraph{Evaluation horizon.}
We use $t^{\max}=1500$ for RoboCasa-Kitchen and DexJoCo,  and $t^{\max}=220/280/300/520$ for
LIBERO-Spatial/Object/Goal/Long, following the default settings.

\section{Experimental Details}
\label{appendix:exp_details}
We collect 100 demonstrations for Pick and Place, 50 each for Stack Cup, Table Sweeping, Air Fryer, and Pump and Scrub, and 320 for Conveyor Package Handling across three embodiments: a single-arm Franka Research 3 with a parallel-jaw gripper (Pick and Place, Stack Cup, Air Fryer), a bimanual OpenArm with two 6-DoF hands (Table Sweeping, Pump and Scrub), and the ALLEX humanoid (Conveyor Package Handling). The Franka setup uses a third-person and a wrist-mounted camera, and the bimanual setup uses a stereo egocentric camera pair, all at 720$\times$1280 without depth; the ALLEX setup uses a head-mounted stereo camera as its sole vision sensor. We use absolute joint positions as the action mode: 8-dim for the Franka tasks, 26-dim for the bimanual tasks, and 48-dim for the ALLEX task, alongside joint torque estimates from all 48 DoF as physical signals. Unless otherwise noted, each model is evaluated over 32 trials.

\subsection{Task Specifications}
\label{appendix:task_details}
\paragraph{Pick and Place} The task requires picking up a single target object from the table and placing it into the plate. To prevent memorization of one grasp pose, we vary the target object across three categories that differ in size, shape, and required grasp width and vary the object position across the workspace to evaluate spatial generalization. We evaluate each model over 32 trials, consisting of 10,11, and 11 trials per object. We use the instruction ``Pick up the [ doll / green cube / blue cup ] and put it into the plate." for this task.

\paragraph{Stack Cup} The task requires two stacking operations in a fixed order within one episode: the left blue cup is stacked on the yellow cup in the middle, and then the right blue cup is stacked on the same yellow cup. Initial cup positions are varied across trials. We evaluate each model over 32 trials. Each successful stack is worth 0.5 points, so a trial that completes only the first stack scores 0.5. We use the instruction ``Stack the left blue cup on the yellow cup in the middle, then stack the right blue cup on the yellow cup in the middle." for this task.

\paragraph{Air Fryer} This task requires sequential multi-stage manipulation with a Franka arm with a gripper: opening the air fryer basket, picking up the food item from the plate and placing it inside the basket, and closing the basket. We evaluate each model over 32 trials. We use the instruction ``Open the air fryer basket, put the food from the plate into the basket, and close the basket.'' for this task.

\paragraph{Table Sweeping} This bimanual task requires the two arms to play asymmetric and simultaneous roles: the left hand holds the dustpan steady against the floor while the right hand sweeps debris into it with a brush. The debris layout is varied across trials. We evaluate each model over 32 trials. A trial receives 0.5 points for grasping the dustpan correctly and another 0.5 points for sweeping the debris into it. We use the instruction ``Hold the dustpan with the left hand and sweep the debris into it with the brush in the right hand" for this task.

\paragraph{Pump and Scrub} This bimanual task requires picking up a scrubber with the left hand and scrubbing the interior of a white bowl, i.e., maintaining sustained periodic contact with a curved surface rather than executing a single grasp-and-place. We evaluate each model over 32 trials. A trial receives 0.5 points for pumping the soap and 0.5 points for scrubbing the bowl. We use the instruction ``Pick up the blue scrubber with the left hand and scrub the white bowl" for this task.

\paragraph{Conveyor Package Handling}
The task requires picking up a parcel from the conveyor belt and placing it on the unloading table. Parcels are fed onto the belt by a human operator, so the robot has to handle each one as it arrives. To prevent memorization of one grasp pose, we vary the parcel across three categories that differ in size, rigidity, and the required grasp type, and vary the parcel position and orientation along the belt to evaluate spatial generalization. We evaluate each model over 5 trials, and each trial consists 10 consecutive packages. We use the instruction ``Bring the package over, orient barcode up, then place it on the conveyor." for this task.

\subsection{Hardware Setup}
\label{appendix:hw_details}
\begin{figure}[htbp]
    \centering
    \begin{subfigure}[b]{0.31\linewidth}
        \centering
        \includegraphics[width=\linewidth]{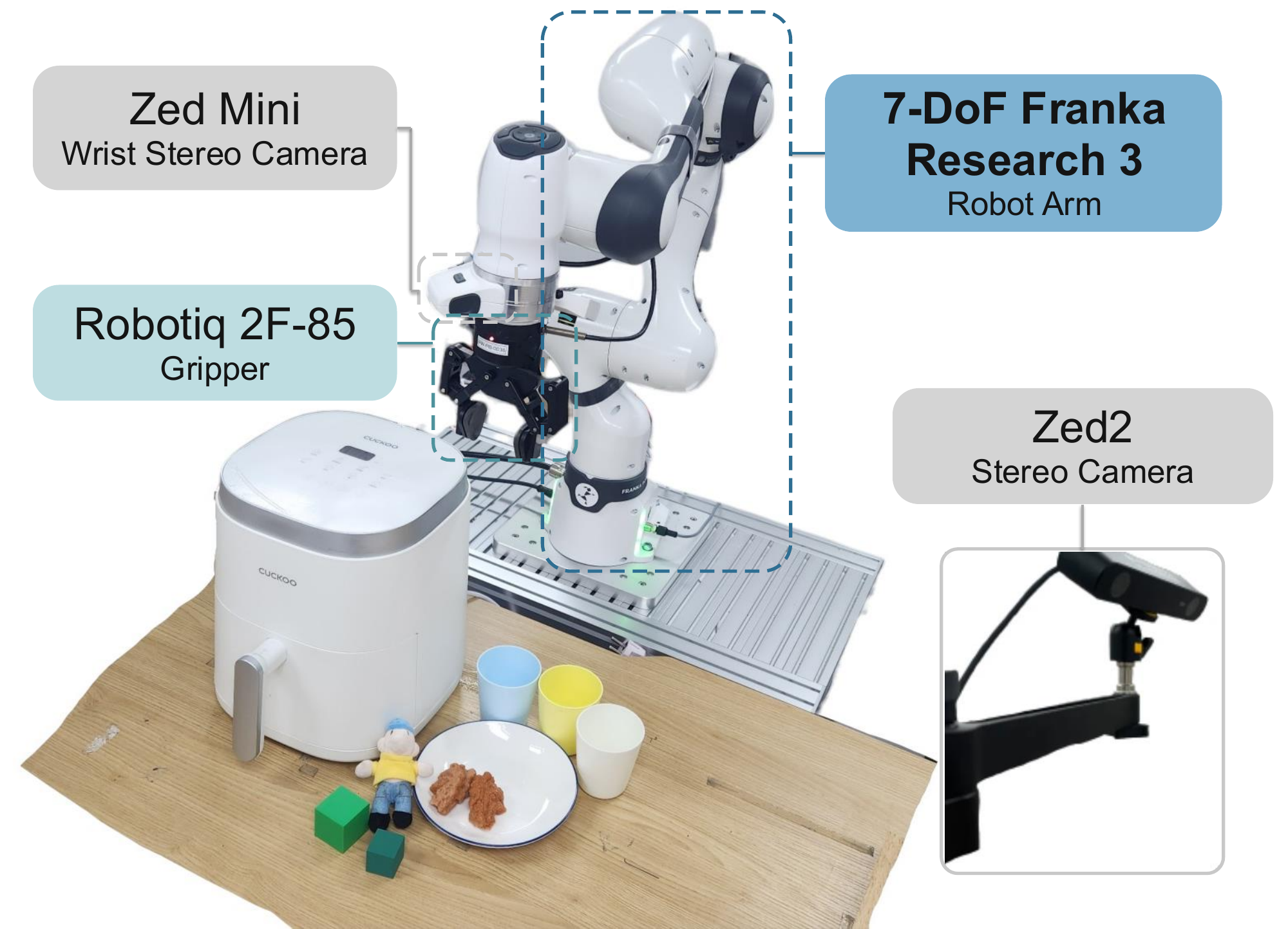} 
        \label{fig:hw_franka}
    \end{subfigure}
    \hfill
    \begin{subfigure}[b]{0.31\linewidth}
        \centering
        \includegraphics[width=\linewidth]{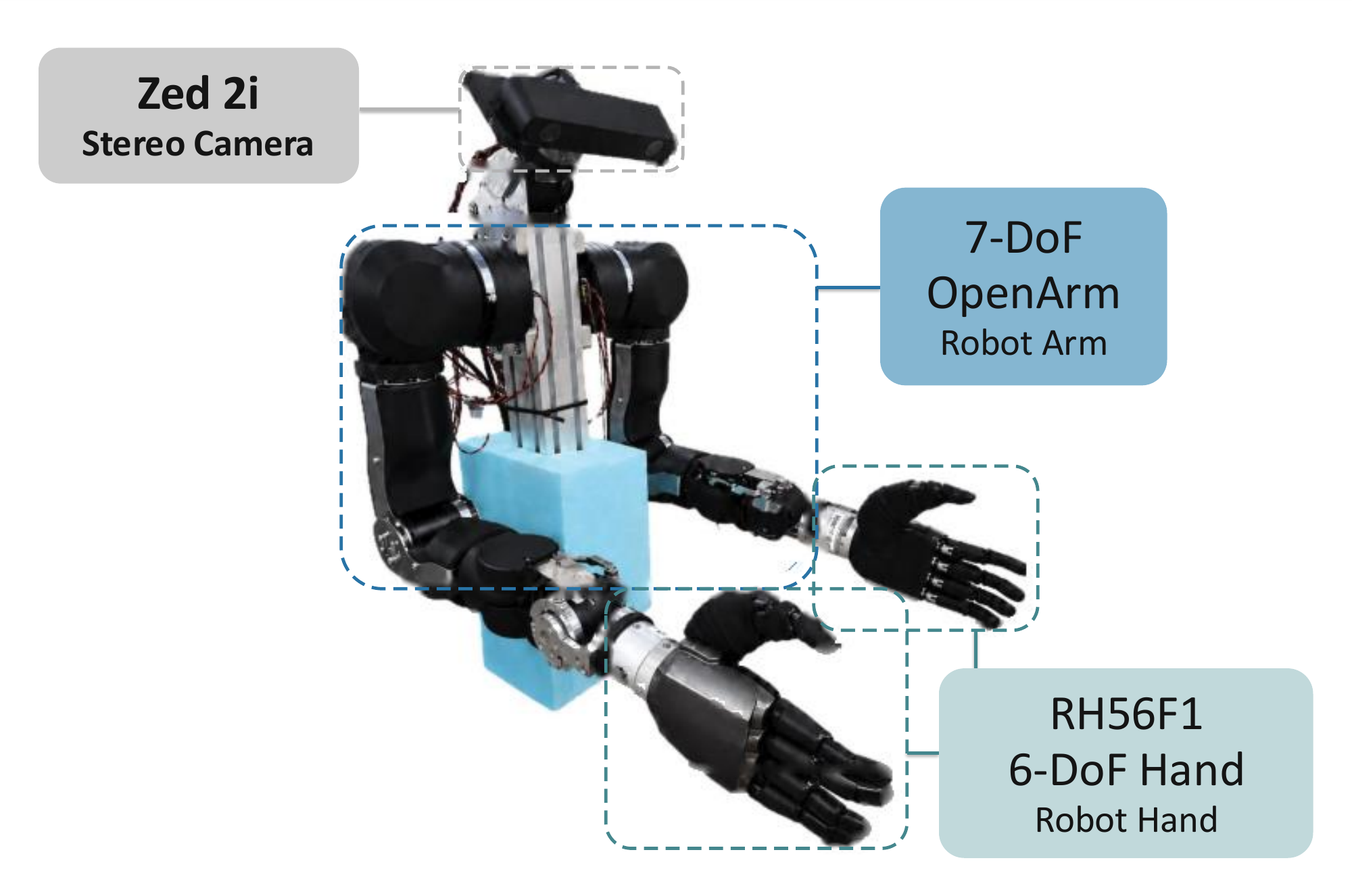}
        \label{fig:hw_openarm}
    \end{subfigure}
    \hfill
    \begin{subfigure}[b]{0.29\linewidth}
        \centering
        \includegraphics[width=\linewidth]{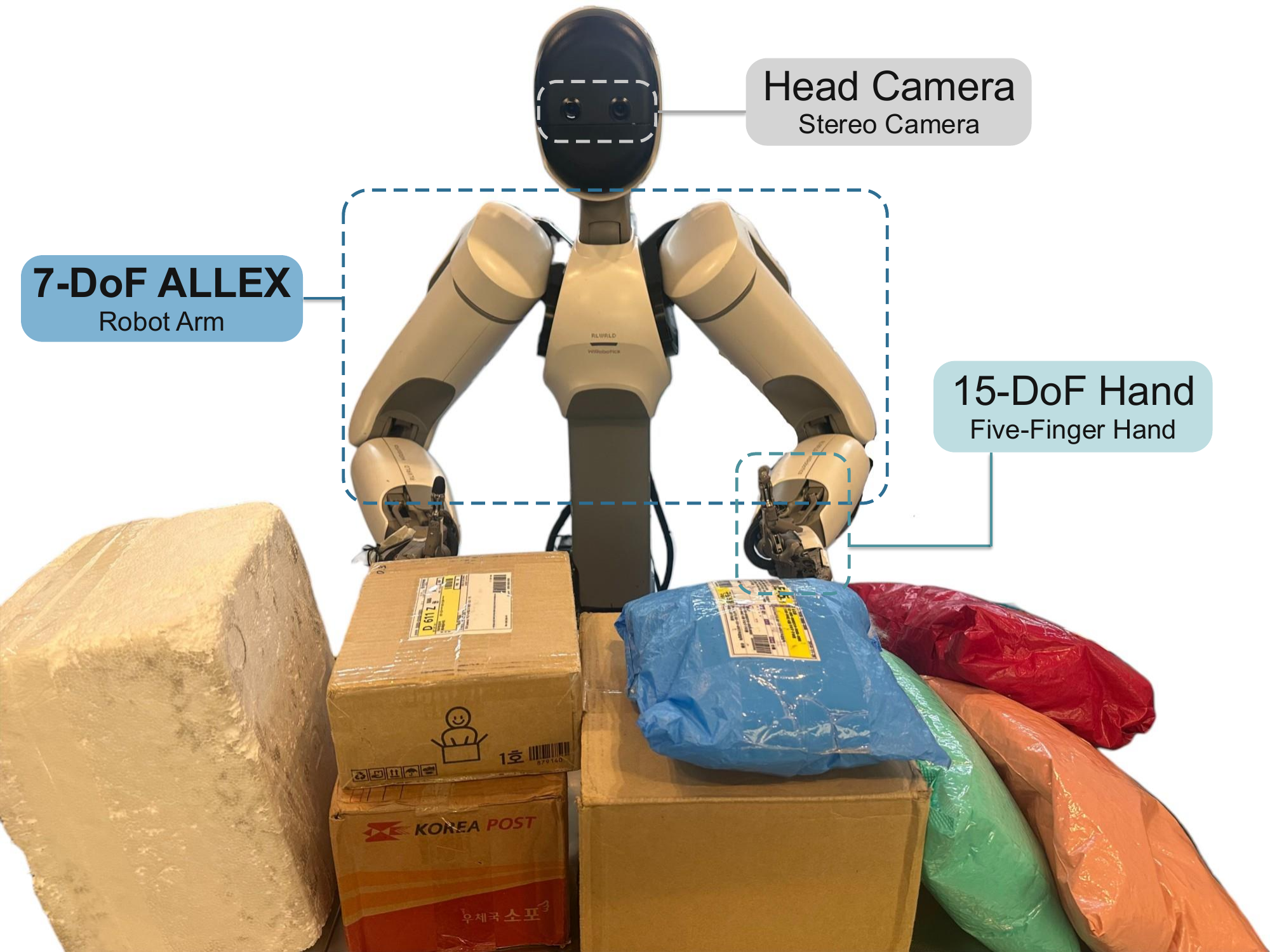}
        \label{fig:hw_allex}
    \end{subfigure}
    
    \caption{Hardware setups and task environments.}
    \label{fig:hardware_setup}
\end{figure}

We evaluate on three platforms: OpenArm, ALLEX, and Franka Research 3, as shown in Figure~\ref{fig:hardware_setup}.

\paragraph{OpenArm.}
OpenArm is a bimanual platform with two 7-DoF arms, each equipped with a 6-DoF Inspire RH56F1 hand. A ZED stereo camera is mounted on the neck, and its left and right views are used as visual inputs. We collect demonstrations by controlling the arms with ExoArm-7, a wearable exoskeleton for OpenArm, and the hands with Manus Pro gloves, where finger joint angles are computed from the fingertip positions via inverse kinematics. The neck is fixed during data collection.

\paragraph{ALLEX.}
ALLEX is a high-DoF upper-body humanoid. We use 48 joints: two 7-DoF arms, two 15-DoF five-finger hands, a 2-DoF waist, and a 2-DoF neck. A head-mounted stereo camera is the only visual input, with no wrist or side cameras. We collect demonstrations using a multi-device teleoperation system with separate interfaces for different robot components. A Meta Quest VR device controls the head and waist, while Vive Trackers provide wrist poses from which arm joint angles are obtained via inverse kinematics. Finger joint angles are similarly recovered from fingertip positions captured by Manus Pro gloves.

\paragraph{Franka Research 3.}
Following DROID~\citep{khazatsky2024droid}, we use a 7-DoF Franka Research 3 arm with a Robotiq 2F-85 gripper. An external ZED camera and a wrist-mounted ZED camera are used as visual inputs. We collect demonstrations via VR teleoperation with a Meta Quest controller.

\subsection{Simulation Hyperparameter Details}
\label{appendix:sim_hyperparams_details}

\paragraph{Training.} We implement \sname{} on two pretrained VLAs, GR00T N1.5 and RLDX-1-IMG,
following their official implementations.
For both models, we freeze the VLM backbone and train the action expert,
the mode-specific decoders, the router, and the judge.
We use $K{=}4$ temporal modes: two fine-grained modes $(16,16)$ and $(8,8)$,
and two quantized modes whose configuration depends on the benchmark
(Table~\ref{tab:hparams_sim_bench}).
We set $\lambda_{\text{sup}}{=}0.1$, $\lambda_{\text{bal}}{=}0.05$, and
$\lambda_{\text{conf}}{=}0.1$, where $\lambda_{\text{sup}}$ and
$\lambda_{\text{bal}}$ increase linearly from zero over the first
5k iterations.
In inference, the quantized group is selected when the judge's predicted
quantizability score is at least the threshold $\delta$.
The remaining hyperparameters are summarized in
Table~\ref{tab:hparams_sim_common}.

\begin{table}[t]
\centering
\small
\caption{Hyperparameter details for training \sname{} on simulation benchmarks.}
\label{tab:hparams_sim_common}
\begin{tabular}{lcc}
\toprule
 & GR00T N1.5 + \sname{} & RLDX-1-IMG + \sname{} \\
\midrule
Optimizer                         & AdamW                              & AdamW \\
Optimizer momentum                & $\beta_1, \beta_2 = 0.95, 0.999$   & $\beta_1, \beta_2 = 0.9, 0.999$ \\
Optimizer weight decay            & 1e-5                               & 1e-5 \\
Learning rate                     & 1e-4                               & 1e-4 \\
Learning rate scheduler           & Cosine decay                       & Cosine decay \\
Warmup iterations                 & 3,000                              & 3,000 \\
Training iterations               & 60,000                             & 60,000 \\
$\lambda_{\text{sup}}, \lambda_{\text{bal}}, \lambda_{\text{conf}}$ & 0.1, 0.05, 0.1 & 0.1, 0.05, 0.1 \\
Router weight $w$ temperature     & 0.5                                & 0.5 \\
Target distribution $r$ temperature & 0.3                              & 0.3 \\
Minimum mode probability          & 0.05                               & 0.05 \\
\bottomrule
\end{tabular}

\vspace{1em}

\caption{Benchmark-specific settings. Quantized modes are given as $(H_k, T_k)$.}
\label{tab:hparams_sim_bench}
\begin{tabular}{lccc}
\toprule
 & RoboCasa-Kitchen & LIBERO & DexJoCo \\
\midrule
Batch size        & 64 & 32 & 64 \\
Quantized modes   & $(16,8),\ (8,4)$ & $(16,9),\ (8,5)$ & $(16,8),\ (8,4)$ \\
\bottomrule
\end{tabular}
\end{table}

\paragraph{Controller configuration.}
\label{appendix:controller_config_sim}
Both RoboCasa-Kitchen and LIBERO use an operational space controller (OSC).
Since temporal action quantization merges consecutive delta actions into a single command, the controller must track a larger displacement per step, and the default position gain ($k_p{=}150$) can undershoot large merged motions. We therefore select $k_p \in \{150, 300, 600\}$ per task, and apply the same per-task gain to all acceleration methods for a fair comparison. For base GR00T N1.5, we use the default gain. 

\FloatBarrier
\subsection{Real Hyperparameter Details}
\label{appendix:Real_hyperparams_details}

\paragraph{Training.}
All real-world experiments use GR00T N1.5 + \sname{} with the same
training hyperparameters as in Table~\ref{tab:hparams_sim_common},
except for the embodiment-specific settings in Table~\ref{tab:hparams_real}.
For ALLEX, we use a longer source horizon, so the fine-grained modes are
$(24,24)$ and $(12,12)$; the other embodiments use $(16,16)$ and $(8,8)$. 
On Franka, Pick-and-Place is trained for 20k iterations, while Air Fryer and Stack Cup are trained for 30k iterations.

\paragraph{Controller configuration.}
\label{appendix:controller_config_real}
On the real robots, temporally quantized commands require both the arm and the end-effector to follow larger changes within a single control step. For the Franka arm, we use Cartesian impedance control with a translational stiffness of 600\,N/m (800\,N/m for stacking), and set the Robotiq 2F-85 to its maximum finger speed (150\,mm/s) and minimum grasp force ($\sim$20\,N). For OpenArm, we use joint PD control with gravity compensation, with shoulder and elbow gains of $K_p \in \{240, 480\}$\,Nm/rad across two configurations.

\begin{table}[h]
\centering
\small
\caption{Embodiment-specific settings for real-world tasks.
Modes are given as $(H_k, T_k)$.}
\label{tab:hparams_real}
\begin{tabular}{lccc}
\toprule
 & Franka & OpenArm & ALLEX \\
\midrule
Batch size               & 64 & 64 & 256 \\
Training iterations      & 30,000 & 40,000 & 30,000 \\
Fine-grained modes       & $(16,16),\ (8,8)$ & $(16,16),\ (8,8)$ & $(24,24),\ (12,12)$ \\
Quantized modes          & $(16,8),\ (8,4)$ & $(16,8),\ (8,4)$ & $(24,12),\ (12,6)$ \\
\bottomrule
\end{tabular}
\end{table}

\FloatBarrier
\subsection{Quantizability Supervision Details}
\label{appendix:quantizability_details}
We design the annotation criteria with the assistance of Claude Opus 5~\citep{anthropic2026opus5}. 
We use per-task success rates under fixed quantization only as task-level cues for constructing the quantizability criteria, and they are never used as supervision targets for the judge.
These task-level cues help reveal quantizability patterns that are difficult to infer from task information alone and guide the construction of the visual checks.
The criterion is decomposed into four to six concrete visual checks.
Each check is rated on a 1--5 scale and contributing either positively or negatively to quantizability (prompts in Appendix~\ref{appendix:prompts}).

We then annotate the full datasets with Gemini 3.8 Flash~\citep{googledeepmind2026gemini38flash}.
For each action chunk, the model receives the first-frame camera observations, task instruction, and computed facts derived from the demonstration, and returns one integer rating per check.
The ratings are combined with fixed weights and normalized to $[0,1]$ to obtain the quantizability score $y$.

\label{appendix:comp_sup_details}

\FloatBarrier
\subsection{Comparison to SAIL}
\label{appendix:adaptation_sail}
We reimplement SAIL~\citep{arachchige2025sail} on the same GR00T-N1.5 and RLDX-1-IMG backbones used in our experiments. We follow the original implementation and hyperparameters, with three changes to keep the comparison fair. First, we parameterize action chunks as end-effector deltas to match the delta control interface of simulation benchmarks. Second, for Error-Adaptive Guidance (EAG), we condition on the four actions of the previous chunk that immediately follow the replanning step, \ie, actions replaced by the new chunk, so that the condition is temporally aligned with the prediction. 
Third, we disable action scheduling.
Instead, SAIL follows the same inference setting as our method and the baselines:
synchronous inference on all tasks except ALLEX conveyor task,
where all methods use asynchronous inference with real-time chunking (RTC)~\citep{black2026real}.
Thus, speedups within each task come from faster motion rather than from the inference setting.

\section{Additional Analysis}

\subsection{Quantitative Analysis on Ground-Truth Data}
\label{appendix:quantitative_gt_analysis}
\noindent\textbf{Setup.}
We examine whether dedicated coarse decoders provide an advantage over directly quantizing the actions predicted by a standard policy.
We evaluate on the RoboCasa-Kitchen simulation benchmark~\citep{nasiriany2024robocasa}, sampling trajectory points across all 24 tasks.
Both \sname{} and the baselines are fine-tuned on the same data and receive the same ground-truth observations.
The two predictions are constructed as follows:
\begin{itemize}[leftmargin=1em,nosep]
    \item \textbf{Ours}: The output of the quantized decoder $D_k$.
    \item \textbf{Base}: The output of decoder, post-processed
    using the same quantization rule as the target of $D_k$.
\end{itemize}
Both are compared against the corresponding ground-truth quantized target
$\mathbf{a}^{(k)}$.

\textbf{Metrics.}
Let
$\hat{\mathbf{a}}=(\hat{\mathbf a}_1,\dots,\hat{\mathbf a}_T)$
and
$\mathbf{a}=(\mathbf a_1,\dots,\mathbf a_T)$
denote the predicted and ground-truth quantized chunks, respectively, with
$\mathbf a_i\in\mathbb{R}^{d}$ over the continuous action dimensions.
We report MSE, cosine similarity, and Dynamic Time Warping (DTW).

\begingroup
\setlength{\abovedisplayskip}{4pt}
\setlength{\belowdisplayskip}{4pt}

MSE measures pointwise magnitude error:
\begin{equation}
\mathrm{MSE}(\hat{\mathbf a},\mathbf a)
=
\frac{1}{Td}
\sum_{i=1}^{T}
\left\|
\hat{\mathbf a}_i-\mathbf a_i
\right\|_2^2 .
\end{equation}

Cosine similarity measures directional agreement:
\begin{equation}
\mathrm{cos}(\hat{\mathbf a},\mathbf a)
=
\frac{
\mathrm{vec}(\hat{\mathbf a})^\top
\mathrm{vec}(\mathbf a)
}{
\left\|\mathrm{vec}(\hat{\mathbf a})\right\|_2
\left\|\mathrm{vec}(\mathbf a)\right\|_2
}.
\end{equation}

DTW measures sequence similarity while allowing temporal misalignment.
With $D(0,0)=0$ and $D(i,0)=D(0,j)=\infty$ for $i,j\geq1$,
\begin{equation}
\begin{aligned}
D(i,j)
&=
\left\|
\hat{\mathbf a}_i-\mathbf a_j
\right\|_2 \\
&\quad +
\min\!\left\{
D(i-1,j),
D(i,j-1),
D(i-1,j-1)
\right\},
\end{aligned}
\end{equation}
and $\mathrm{DTW}(\hat{\mathbf a},\mathbf a)=D(T,T)$.
\endgroup

\textbf{Results.}
Table~\ref{tab:dtw_compare} shows that the dedicated coarse decoders
consistently outperform post-hoc quantization of the baseline predictions
across both configurations and all three metrics.
This indicates that directly learning the quantized action target better
preserves the ground-truth motion than quantizing a fine-grained prediction
after inference.

\begin{table}[t]
\centering
\caption{\textbf{Quantized-action prediction accuracy on RoboCasa-Kitchen.}
\emph{Ours} uses dedicated coarse decoders, while \emph{Base} quantizes
GR00T~N1.5 predictions using the same SO(3)-aware rule.
$(H,q)$ denotes the source horizon and quantization level, and \emph{Both}
averages the two configurations.}
\label{tab:dtw_compare}
\vspace{0.1cm}

\begin{tabular}{llcc}
\toprule
Decoder $(H,q)$ & Metric & \emph{Ours} & \emph{Base} \\
\midrule
\multirow{3}{*}{\textsc{(16,2)}}
& MSE~$\downarrow$    & \textbf{0.0273} & 0.0344 \\
& DTW~$\downarrow$    & \textbf{3.209}  & 3.777 \\
& Cosine~$\uparrow$   & \textbf{0.850}  & 0.809 \\
\midrule
\multirow{3}{*}{\textsc{(8,2)}}
& MSE~$\downarrow$    & \textbf{0.0163} & 0.0243 \\
& DTW~$\downarrow$    & \textbf{1.359}  & 1.658 \\
& Cosine~$\uparrow$   & \textbf{0.886}  & 0.847 \\
\midrule
\multirow{3}{*}{Both}
& MSE~$\downarrow$    & \textbf{0.0218} & 0.0294 \\
& DTW~$\downarrow$    & \textbf{2.284}  & 2.718 \\
& Cosine~$\uparrow$   & \textbf{0.868}  & 0.828 \\
\bottomrule
\end{tabular}
\end{table}

\subsection{Rollout Trajectory Comparison}
\begin{figure}[H]
  \centering
  \includegraphics[width=\linewidth]{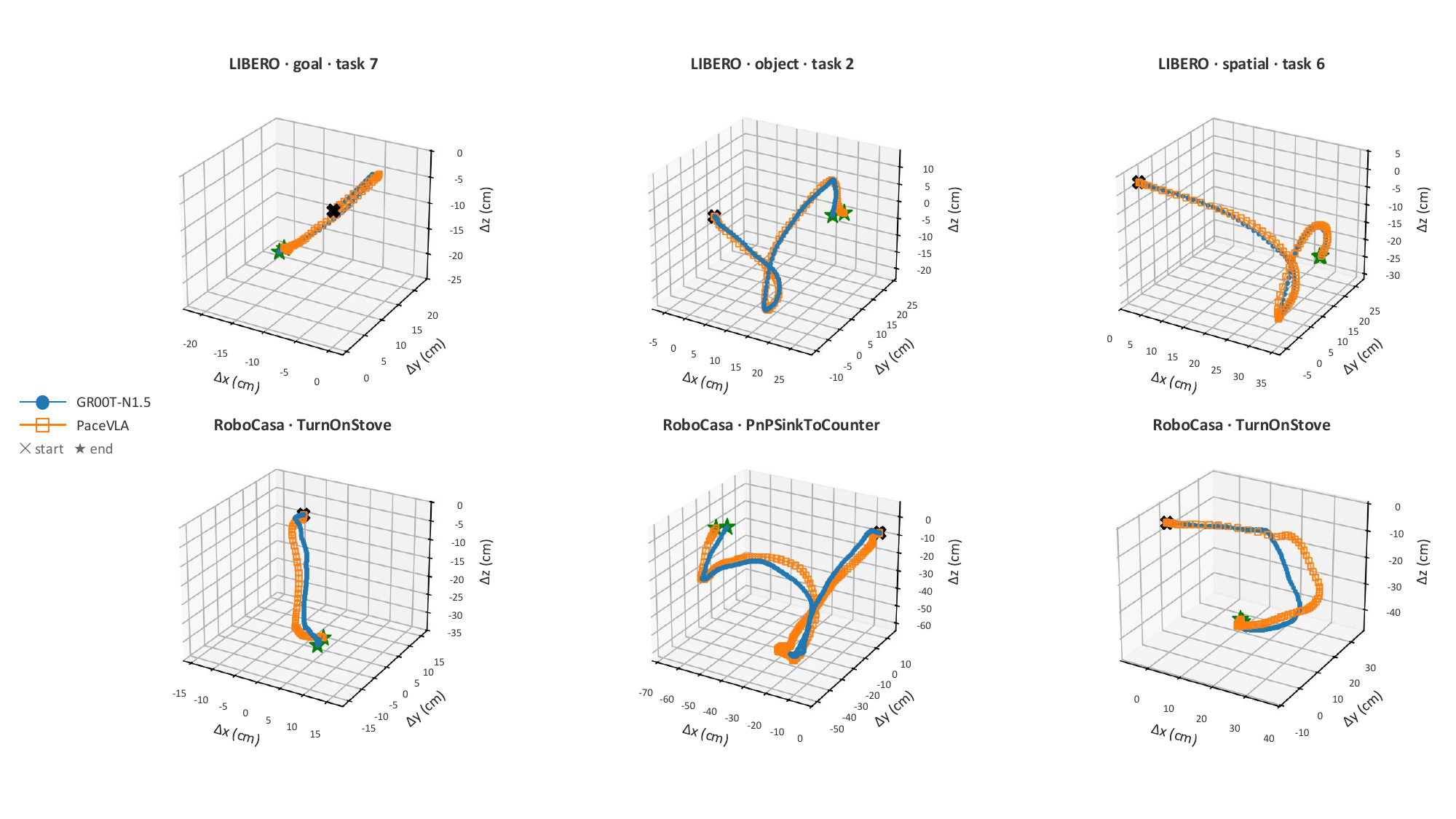}
  \caption{\textbf{End-effector rollout trajectories.}  End-effector trajectories of the base policy (blue) and \sname (orange) from the same initial state; \sname traverses the same path with fewer, sparser steps.}
  \label{fig:rollout_traj_comp2}
\end{figure}

\subsection{Failure Case Analysis}

\begin{figure}[htbp]
    \centering
    \includegraphics[width=\linewidth]{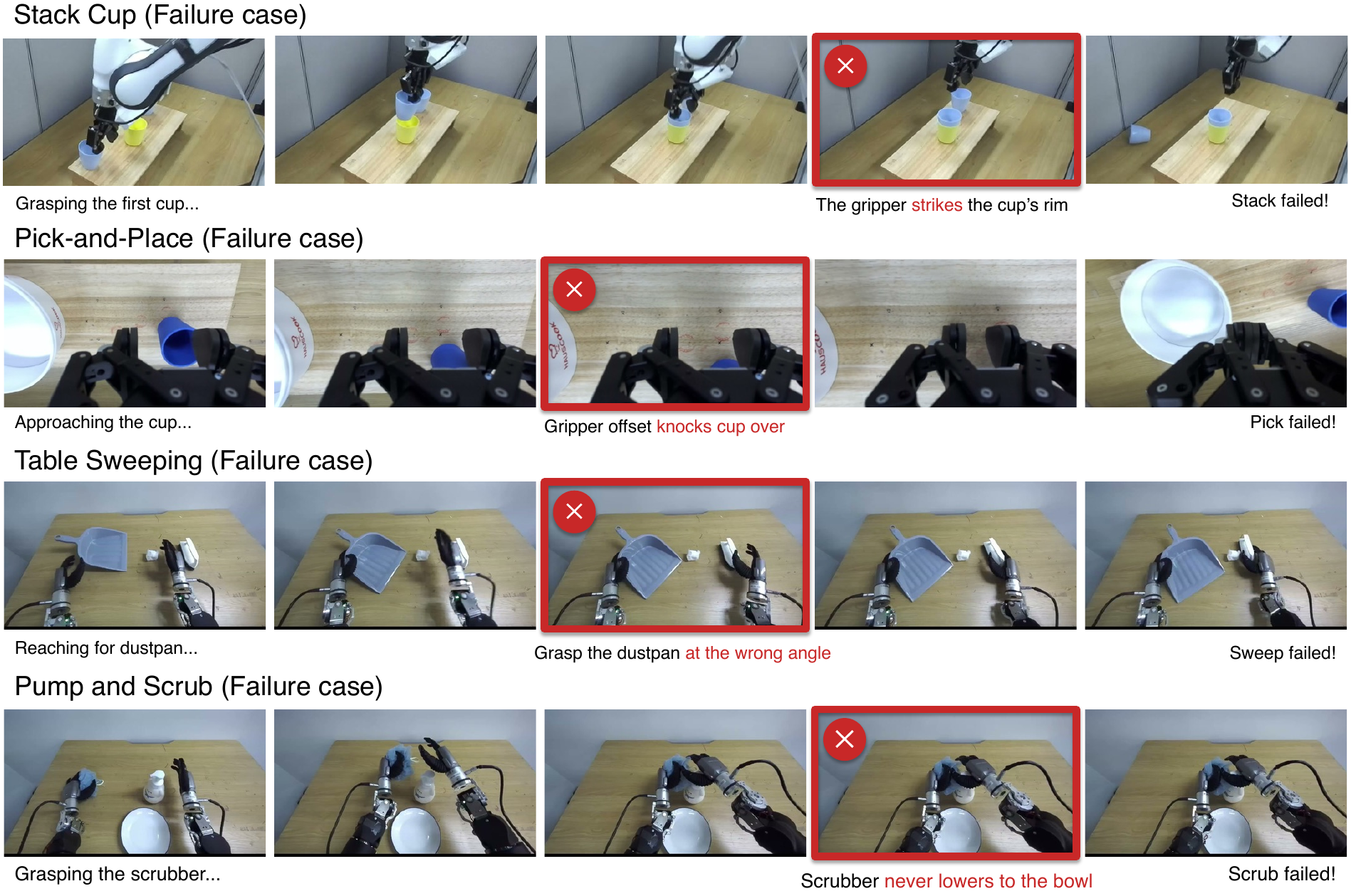}
    \caption{Failure cases of \sname{}. Most failures come from small targeting errors (Pick-and-Place, Stack Cup) or from the hands and tools blocking the contact region in the egocentric view (Table Sweeping, Pump and Scrub).}
    \label{fig:failure_cases}
\end{figure}

Across the failed rollouts we inspected (Figure~\ref{fig:failure_cases}), most failures seem to come from perception rather than from the temporal decision. All four decoders use the same visual and language features from the backbone, so if the scene is misread, the choice of decoder makes little difference. In Pick-and-Place, most failures happen during grasping. The judge assigns low quantizability in this phase and the router switches to fine-grained modes, but the trajectory is aimed slightly off the object, so the gripper knocks the cup over and closes on air, or holds the object loosely and drops it during transport. In Stack Cup, the gripper misses the first cup slightly and hits its edge, so the first stack fails and the robot cannot recover. The base model fails in the same way, so we believe these errors are mostly due to perception rather than temporal adaptation. 

We also see a similar pattern in the OpenArm bimanual hand tasks. The only visual input is an egocentric stereo camera, and the hands and tools often block the view of the contact region. In Table Sweeping, the left hand grasps the dustpan at the wrong angle, so its lip does not sit flush against the table, and the debris swept by the right hand piles up against the lip instead of going in. In Pump and Scrub, the policy pumps soap onto the scrubber but never proceeds to scrubbing, possibly because the hands occlude the scrubber from the camera, so the policy cannot see that the soap has already been applied. In these cases the missing information is in the observation, not in the trajectory, so finer temporal granularity is unlikely to help. These cases suggest that \sname{} is still limited by the quality of its perception, and that better sensing, such as depth or tactile input, would add to the gains from adaptive temporal execution.

\clearpage
\section{Prompts used on Dataset Annotation}
\label{appendix:prompts}
\definecolor{promptrule}{HTML}{3D5A80}
\definecolor{promptback}{HTML}{F7F9FB}

\newtcolorbox{promptbox}[2][]{%
  enhanced,
  sharp corners,
  colback=promptback,
  colframe=promptrule,
  boxrule=0.6pt,
  left=5pt,
  right=5pt,
  top=7pt,
  bottom=4pt,
  before skip=3pt,
  after skip=3pt,
  fonttitle=\small\bfseries,
  coltitle=white,
  title={#2},
  attach boxed title to top left={
    xshift=6pt,
    yshift=-2.2pt
  },
  boxed title style={
    colback=promptrule,
    colframe=promptrule,
    sharp corners,
    boxrule=0pt,
    left=4pt,
    right=4pt,
    top=1.5pt,
    bottom=1.5pt
  },
  #1
}

\fvset{
  fontsize=\scriptsize,
  fontfamily=tt,
  breaklines=true,
  breakanywhere=false,
  breakindent=0pt,
  xleftmargin=0pt,
  baselinestretch=0.90
}

\begin{promptbox}[before skip=0pt]{Sim: RoboCasa-Kitchen}
\begin{Verbatim}
[USER]
<3 images>
You are judging one moment of a kitchen robot's motion, to decide whether the arm could get through the stretch of motion ahead of it faster -- reaching the same places in fewer and longer steps -- without changing the outcome.

Going faster leaves the arm fewer chances to correct itself along the way. That costs most where the arm still has to arrive somewhere exactly: setting down what it is carrying onto the spot it must end up on, or closing the gripper on one exact point -- and, when it is the gripper closing, more still if that spot is hemmed in. It costs least where the arm only acts on something that is anchored in place -- a button or switch pressed, a knob or tap turned, a door pushed shut. There a longer step just presses or turns a little harder and nothing is lost.

Judge the moment in front of you, not the task as a whole. One task passes through both kinds from one moment to the next.

You are shown 3 camera views: agentview-left, agentview-right, and a wrist (eye-in-hand) close-up. The wrist camera is mounted on the gripper, so objects normally look close in it -- general closeness is normal. Use the wrist view to judge the close moments: the gripper actually closing on something, and something being set down onto the spot it must end up on.

The robot was told: {instruction}

{computed facts}

The measurements above are stated as fact -- do not re-estimate or repeat them. Answer each check from what the cameras show about the MOMENT in front of you, read together with those measurements.
Answer each check on its own line as "A) 3", in order, nothing else -- one digit from 1 to 5 per check, rating how far that check describes this moment:
  5 = clearly true of this moment
  4 = mostly true
  3 = partly true
  2 = barely true
  1 = not true at all
A grade refers only to the check on that line.
A) Is the gripper CLOSING on its target right now, or about to -- the fingers
   already around the handle, rim or object and only the squeeze left, so that a
   slip here loses it?
B) Is that closing spot hemmed in -- reached into a cabinet, a microwave, an oven or
   a sink, or lined up with a slot -- so the gripper has little room around it as
   it closes?
C) Is the arm just pressing, turning or pushing something that is anchored in
   place -- a button or switch, a knob or tap, a door or drawer being shoved shut
   -- and not something it has to keep hold of while it travels, the way a handle
   is held on to and pulled to swing a door or drawer open?
D) Is the object ALREADY RESTING where it is meant to end up, taking its own
   weight, with nothing left but to open the hand and draw back?
E) Is the gripper carrying something through open space -- already holding it,
   still well away from wherever it is going?
F) Is the arm STILL LOWERING the object onto its target right now -- the gripper
   not yet opened, the object not yet taking its own weight, coming down onto
   the spot it must end up on?
Answer:
\end{Verbatim}
\end{promptbox}

\clearpage

\begin{promptbox}[before skip=0pt]{Sim: LIBERO}
\begin{Verbatim}
[USER]
<2 images>
You are judging one moment of a table-top robot's motion, to decide whether the arm could get through the stretch of motion ahead of it faster -- reaching the same places in fewer and longer steps -- without changing the outcome.

Going faster leaves the arm fewer chances to correct itself along the way. That costs most when the gripper has to close on one exact thing -- and more still when the way in is threaded: the object perched on top of another, or wedged in among things standing close by, so that there is one line in and a longer step swings wider than that line. It costs least where whatever is being moved only has to land somewhere that will catch it, and at the moments either side of the work -- crossing open space, carrying something with room all round it, drawing back after letting go.

Judge the moment in front of you, not the task as a whole. One task passes through both kinds from one moment to the next: the same episode that threads a bowl out from between two dishes then carries it through open air.

You are shown 2 camera views of this one moment: a scene view and a wrist (eye-in-hand) close-up. The wrist camera is mounted on the gripper, so objects normally look close in it -- general closeness is normal. Use the wrist view to judge the close moments: the gripper actually closing on something, and something being set down into the place it must end up.

The robot was told: {instruction}

{computed facts}

The measurements above are stated as fact -- do not re-estimate or repeat them. Answer each check from what the cameras show about the MOMENT in front of you, read together with those measurements.
Answer each check on its own line as "A) 3", in order, nothing else -- one digit from 1 to 5 per check, rating how far that check describes this moment:
  5 = clearly true of this moment
  4 = mostly true
  3 = partly true
  2 = barely true
  1 = not true at all
A grade refers only to the check on that line.
A) Is the place this object has to end up ROOMY FOR IT -- a wide bowl, an open
   basket, a broad tray or an open stretch of table with slack all round -- rather
   than a plate, a rack, or a spot barely bigger than the thing itself?
B) Is the gripper closing on the object right now, or about to -- the fingers
   already around it and only the squeeze left, so that a slip here loses it?
C) Is that closing spot threaded -- the thing being taken perched on top of another
   object, or wedged in among things standing close by -- so that there is only one
   line in for the hand?
D) Is the hand ALREADY OPEN and coming away, the object left sitting where it was
   meant to end up and holding its own weight without the fingers?
E) Is this a job that needs no hold kept on anything -- a push, a knob turned, a
   drawer pressed shut -- so that once it is sent it carries on without the hand?
Answer:
\end{Verbatim}
\end{promptbox}

\begin{promptbox}[before skip=0pt]{Sim: DexJoCo}
\begin{Verbatim}
[USER]
[IMAGE]
[IMAGE]
You are judging one moment of a robot's motion with a dexterous hand, to decide whether the arm could get through the stretch of motion ahead of it faster -- reaching the same places in fewer and longer steps -- without changing the outcome.

Going faster leaves the arm fewer chances to correct itself along the way. Changing the working configuration of a held object and directing a held tool onto a localized target depend on the relation between the fingers, object and target. Maintaining a supported press and carrying an already grasped object through clear space impose different constraints. Use the visible interaction to judge these checks; do not infer contact or a secure grasp from commanded motion alone.

Judge the moment in front of you, not the task as a whole. One task passes through different interactions from one moment to the next.

You are shown 2 camera views of this one moment: image 1 is the scene view (observation.images.ego_right), and image 2 is a wrist (eye-in-hand) close-up (observation.images.wrist). The wrist camera is mounted on the hand, so objects normally look close in it -- general closeness is normal. Use the wrist view to judge the close moments: the fingers operating the held object, a tool approaching its target, and a finger pressing a supported target.

The robot was told: {instruction}

{computed facts}

The measurements above are stated as fact -- do not re-estimate or repeat them. Answer each check from what the cameras show about the MOMENT in front of you, read together with those measurements.
Answer each check on its own line as "A) 3", in order, nothing else -- one digit from 1 to 5 per check, rating how far that check describes this moment:
  5 = clearly true of this moment
  4 = mostly true
  3 = partly true
  2 = barely true
  1 = not true at all
A grade refers only to the check on that line.
A) Are the fingers operating movable parts of the object being held—closing
   jaws or folding an articulated part—so that changing the relation between
   those parts is the operation, rather than merely holding the object?
B) Is a held tool being directed into an impact on a localized target, rather
   than making a sustained supported press? Use visible held-tool/target geometry together with the command motion to judge
   preparation for or execution of a targeted strike. Command targets do not establish
   that an impact actually occurred; the presence of a hammer alone is not enough.
C) Is a finger or hand pressing a target supported by the surrounding surface,
   without lifting it, striking it with a tool, or manipulating movable parts
   within a grasp? A surface supports the target; do not assume it is bolted down.
D) Is an already grasped object being moved or tilted as a whole through clear
   space, without changing the working configuration of its movable parts,
   acquiring/releasing the grasp, or entering a visibly tight target region?

Answer:
\end{Verbatim}
\end{promptbox}

\begin{promptbox}{Real-World: Franka+Gripper Tasks}
\begin{Verbatim}
[USER]
<2 images>
You are judging one moment of a table-top robot's motion, to decide whether the arm could get through the stretch of motion ahead of it faster -- reaching the same places in fewer and longer steps -- without changing the outcome.

Going faster leaves the arm fewer chances to correct itself along the way. That costs most at the two moments where the object can be lost: when the hand is closing on it, and when it is being brought down into the place it has to end up. Both get worse the less room there is -- letting a cube down into a cup barely wider than itself leaves no slack, while letting the same cube down onto an open plate does.

Most moments are neither. Crossing empty space to reach for something, carrying something with room all round it, or drawing the hand back after letting go -- a longer step changes nothing there. That is the common case.

Judge the moment in front of you, not the task as a whole. One task passes through both kinds from one moment to the next.

You are shown 2 camera views of this one moment: a scene view and a wrist (eye-in-hand) close-up. The wrist camera is mounted on the gripper, so objects normally look close in it -- general closeness is normal. Use the wrist view to judge the close moments: the hand actually closing on the object, and the object being brought down into the place it must end up.

The robot was told: {instruction}

{computed facts}

The measurements above are stated as fact -- do not re-estimate or repeat them. Answer each check from what the cameras show about the MOMENT in front of you, read together with those measurements.
Answer each check on its own line as "A) 3", in order, nothing else -- one digit from 1 to 5 per check, rating how far that check describes this moment:
  5 = clearly true of this moment
  4 = mostly true
  3 = partly true
  2 = barely true
  1 = not true at all
A grade refers only to the check on that line.
A) Is the hand closing on the object right now, or about to -- the fingers
   already around it and only the squeeze left, so that a slip here loses it?
B) Is the place this object has to end up ROOMY FOR IT -- a broad plate or an
   open dish with slack all round -- rather than a cup or a container barely
   wider than the thing itself?
C) Is the hand ALREADY OPEN and coming away, the object left sitting where it
   was meant to end up and holding its own weight without the fingers?
D) Is the arm STILL LOWERING the object onto its target right now -- the hand
   not yet opened, the object not yet taking its own weight, coming down onto
   the spot it must end up on?
Answer:
\end{Verbatim}
\end{promptbox}

\begin{promptbox}{Real-World: OpenArm+Bimanual Hand Tasks}
\begin{Verbatim}
[USER]
<2 images>
You are judging one moment of a two-armed robot's motion, each arm ending in a dexterous hand, to decide whether it could get through the stretch of motion ahead faster -- reaching the same places in fewer and longer steps -- without changing the outcome.

Going faster leaves the arms fewer chances to correct themselves along the way. That costs most where a hand has to arrive somewhere exactly or close on something exactly: taking hold of a tool, pushing a hand-held thing down onto something that is only standing on the table and can tip or slide, and guiding what is held into a small target region. It costs least where a hand is crossing open space, carrying something that is already firmly in the grasp with room all round it.

The two hands may be doing different things at the same moment -- one may hold something steady while the other works. Read both. Use the visible interaction; do not infer contact or a secure grasp from commanded motion alone.

Judge the moment in front of you, not the task as a whole. One task passes through different interactions from one moment to the next.

You are shown 2 camera views of this one moment, both scene views of the same workspace from slightly different angles (image 1 is observation.images.ego_left, image 2 is observation.images.ego_right). There is no wrist camera; neither view is an eye-in-hand close-up. This robot has two arms, each with a dexterous hand, and they may be doing different things at the same time -- read both hands.

The robot was told: {instruction}

{computed facts}

The measurements above are stated as fact -- do not re-estimate or repeat them. Answer each check from what the cameras show about the MOMENT in front of you, read together with those measurements.
Answer each check on its own line as "A) 3", in order, nothing else -- one digit from 1 to 5 per check, rating how far that check describes this moment:
  5 = clearly true of this moment
  4 = mostly true
  3 = partly true
  2 = barely true
  1 = not true at all
A grade refers only to the check on that line.
A) Is a hand closing on a tool or object to take hold of it right now, or about
   to -- the fingers already around it and only the squeeze left, so that a slip
   here loses it -- rather than already holding it securely?
B) Is a hand pushing down on something that is only standing on the table and is
   not held flat by it -- a bottle, a pump head, an upright container -- so that
   going too far tips it over or slides off it, rather than pressing on something
   the table holds down?
C) Is what a hand holds being guided into or over a small target region -- across
   a rim or lip, into a narrow opening, onto a spot only a little larger than the
   thing itself -- so that missing that line loses the attempt?
D) Is an already grasped object being moved or tilted as a whole through clear
   space, without changing the working configuration of its movable parts,
   acquiring/releasing the grasp, or entering a visibly tight target region?
E) Is a hand moving through open space touching nothing at all -- on its way
   somewhere, or drawing back after letting go?

Answer:
\end{Verbatim}
\end{promptbox}

\begin{promptbox}{Real-World: Allex Conveyer Package Handling Task}
\begin{Verbatim}
[USER]
<2 images>
You are judging one instant of a two-armed robot with hands, to decide how much of the next stretch of its motion could be thinned out -- how many of its commanded poses could be dropped, letting the arms travel further between the ones that remain, without changing the outcome.

What thinning takes away is the arms' chance to correct on the way, and there are two ways that loses the object. One is a hold that exists only as a balance: the thing is trapped between two hands and nothing is closed around it, so the hold IS the push of one hand against the other. If that thing is stiff it cannot give, and a coarser swing changes the balance and drops it. Something soft squashes and stays pinched.

The other is a load that does not hold its own shape. A bag or a sagging thing hangs, swings and shifts while it is carried, and the further the arms travel between commanded poses the further it swings.

Most moments are neither. Sliding something across to the other side, pushing something that is still resting on a surface, or moving with nothing in the hands -- a longer step changes nothing.

Judge the moment in front of you, not the name of the job. One segment passes through several of these from one second to the next.

The two images are the left and right camera views of this one moment.

The robot was told: {instruction}

{computed facts}

The measurements above are stated as fact -- do not re-estimate or repeat them. Answer each check from what the cameras show about the MOMENT in front of you, read together with those measurements.
Answer each check on its own line as "A) 3", in order, nothing else -- one digit from 1 to 5 per check, rating how far that check describes this moment:
  5 = clearly true of this moment
  4 = mostly true
  3 = partly true
  2 = barely true
  1 = not true at all
A grade refers only to the check on that line.
A) Are the two hands SQUEEZING something between them to hold it, and is that
   thing HARD -- a box, a carton, something that will not squash?
B) Is the item a POUCH OR PACKET -- a soft mailer, a padded envelope, a
   plastic parcel -- rather than a rigid carton with square sides?
C) Is the thing only being SENT ACROSS -- slid or passed over to the other side,
   or shoved on its way -- with no particular spot it has to come to rest on?
D) Are the hands ON THEIR WAY -- reaching out towards something, or drawing back
   from what they have finished, carrying nothing between them?
Answer:
\end{Verbatim}
\end{promptbox}

\end{document}